\pdfoutput=1

\documentclass[11pt]{article}

\usepackage[final]{ACL2023}

\usepackage{algorithm}
\usepackage{algorithmic}
\usepackage{times}
\usepackage{latexsym}
\usepackage{amsmath}
\usepackage{lscape} 

\usepackage[T1]{fontenc}

\usepackage[utf8]{inputenc}

\usepackage{microtype}

\usepackage{inconsolata}
\usepackage{xspace}
\usepackage{graphicx}
\usepackage{booktabs}
\usepackage{dashrule}
\usepackage{xcolor,colortbl}
\usepackage{enumitem}
\usepackage{mdframed}
\usepackage{relsize}
\usepackage{enumitem}
\usepackage[hang,flushmargin]{footmisc}

\definecolor{paleaqua}{rgb}{0.74, 0.83, 0.9}

\title{\mirage: Automatic Multilingual Benchmark Arena for Retrieval-Augmented Generation Systems}

\author{
Nandan Thakur$^1$, Suleman Kazi$^{2}$, Ge Luo$^2$, Jimmy Lin$^1$, Amin Ahmad$^2$
\\[1ex]
$^1$~University of Waterloo, Canada \quad \quad $^2$~Vectara, USA \\[1ex]
        \texttt{\{nandan.thakur,jimmylin\}@uwaterloo.ca} \\
        \texttt{\{suleman,rogger,amin\}@vectara.com}\\
}

\newcommand{\custom}[1]{\textsc{\normalsize #1}}
\newcommand{\customsize}[1]{\textsc{#1}}
\newcommand{\mirage}[0]{\customsize{Mirage}-\customsize{Bench}\xspace}
\newcommand{\miracl}[0]{\custom{miracl}\xspace}

\begin{document}
\maketitle
\begin{abstract}

Traditional retrieval-augmented generation (RAG) benchmarks evaluate systems using heuristic-based metrics, but these require human preferences as the ground truth for reference. In contrast, arena-based benchmarks, where systems compete against each other, require an expensive large language model (LLM) as a judge for a reliable evaluation. 
We present a simple efficient technique to combine the best of both worlds. 
The idea is to train a surrogate judge using heuristic metrics as input, to output the LLM as a judge prediction.
In our work, we develop \mirage,\footnote{
\mirage has been coined as {\footnotesize (\textsc{\underline{Mi}racl}~+~\textsc{\underline{rag}} +~\textsc{\underline{e}valuation} + \textsc{\underline{Bench}mark)}}.} a synthetic arena-based RAG benchmark for 18 diverse languages on Wikipedia focused on multilingual answer generation evaluation. 
It extensively couples both heuristic features and LLM as a judge for evaluation. 
We benchmark 19 multilingual LLMs, and observe a high correlation (Kendall Tau ($\tau$) = 0.909) using our surrogate judge 
and between GPT-4o as a teacher using the Bradley-Terry framework. Our results show proprietary and large open-source LLMs currently dominate on \mirage. Our code and datasets are made publicly available here: \url{https://github.com/vectara/mirage-bench}.
\end{abstract}

\begin{figure}[t]
\centering
\begin{center}
    \includegraphics[trim=0 22 0 22,clip,width=0.48\textwidth]{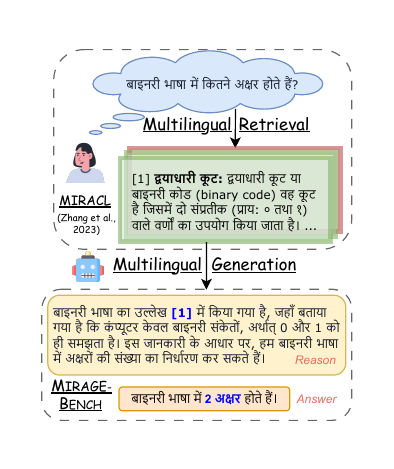}
    \caption{Multilingual naive RAG pipeline in Hindi (hn). In \mirage, we reuse the oracle retrieval set (query and oracle judged passages) from \miracl~\cite{zhang:2023} and focus on evaluating the answer generation stage with multilingual LLMs.}
    \label{fig:multilingual-rag}
\end{center}
\vspace*{-\baselineskip}
\end{figure}

\begin{figure*}[t]
\centering
\begin{center}
    \includegraphics[trim=0 0 0 0,clip,width=\textwidth]{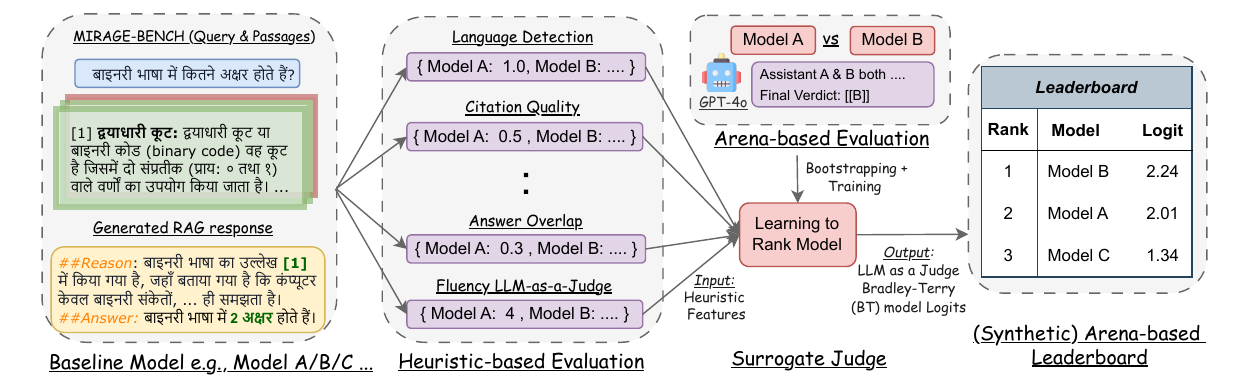}
    \caption{The \mirage evaluation flowchart consists of three steps: (i) heuristic-based features evaluating the baseline model response across several dimensions; (ii) exhaustive pairwise comparisons with GPT-4o as a judge on a small subset of queries to train our surrogate judge. (iii) After training, we utilize our surrogate judge to output the model ranking on the whole subset of queries, to construct the synthetic RAG arena-based leaderboard.}
    \label{fig:MIRAGE-flowchart}
\end{center}
\vspace*{-\baselineskip}
\end{figure*}

\vspace{-2mm}
\section{Introduction}

Large language models (LLMs) have recently gained popularity for information-seeking queries leading to the widespread adoption of retrieval-augmented generation (RAG) \cite{guu:2020, lewis:2020, izacard:2021, borgeaud:2022}. 
The naive RAG setup traditionally includes a retrieval and a generation stage, conducted sequentially. 
RAG systems such as Bing Search \cite{bingsearch} provide grounded responses, i.e., statements that include citations to one or more retrieved passages. 
The citations reduce factual hallucinations with easy verifiability and improve faithfulness to passages provided within context \cite{khandelwal:2020, lewis:2020, gao:2023, gao:2023b, liu:2024}. 
However, existing RAG benchmarks are English-centric, due to uneven and scarce data available across multiple languages \cite{thakur:2024}. So far, it is unclear how existing LLMs perform in multilingual RAG, i.e., where queries and passages are non-English and the LLM generates the answer in the same language. 
An example of a naive RAG pipeline in Hindi (hn) is shown in \autoref{fig:multilingual-rag}.

RAG benchmarks can be broadly classified as either (i) \emph{heuristic-based}, where benchmarks handcraft evaluation metrics (e.g., faithfulness or fluency) to evaluate systems on multiple dimensions \cite{gao:2023, chen:2024, yang:2024} or (ii) \emph{arena-based}: where systems compete each other in a tournament and an LLM-based teacher is used as the judge for evaluation \cite{rackauckas:2024, pradeep:2024}. 
Heuristic-based benchmarks are computationally cheaper to evaluate but require human preferences as gold truth for reference. 
They also face challenges in aggregating different features into a ranking order for models.
On the other hand, arena-based benchmarks require a high-performance or reasoning-intensive LLM as a teacher for accuracy \cite{zheng:2024}, which makes exhaustive pairwise comparisons expensive for a large set of models. For instance, evaluating a single query on 19 models involves $\binom{19}{2}=171$ comparisons and costs between \$5--10 USD with GPT-4o \cite{openai:2023}.

In our work, we get the best of both worlds with a surrogate judge, a learning to rank model, e.g. random forest \cite{kamho:1995}, using heuristic features to estimate an arena-based leaderboard obtained with a Bradley-Terry model \cite{hunter:2004} from pairwise evaluations using an LLM as a judge \cite{zheng:2024}. 
We use bootstrapping to obtain confidence bounds for better statistical estimates. 
After training, the learning to rank model can be used to estimate the performance of newer released models reliably in the future without the expensive LLM as a judge. It also provides better interpretability and is easily retrainable with a different or newer set of heuristic features.

\begin{table*}[t]
\resizebox{\textwidth}{!}{
\centering
\begin{tabular}{lrrrrrrrrrrrrrrrrrr}
\toprule
 & \textbf{ar} & \textbf{bn} & \textbf{de} & \textbf{en} & \textbf{es} & \textbf{fa} & \textbf{fi} & \textbf{fr} & \textbf{hi} & \textbf{id} & \textbf{ja} & \textbf{ko} & \textbf{ru} & \textbf{sw} & \textbf{te} & \textbf{th} & \textbf{yo} & \textbf{zh} \\
\midrule
\multicolumn{19}{c}{\textbf{\mirage Evaluation Dataset}} \\
\midrule
\textbf{\#~Queries} & 1501 & 411 & 304 & 787 & 617 & 632 & 1183 & 343 & 350 & 939 & 797 & 213 & 1247 & 481 & 150 & 730 & 119 & 391 \\
\textbf{\# Avg. Query Tokens} & 10.2 & 17.6 & 10.4 & 9.1 & 11.4 & 14.1 & 12.2 & 10.4 & 17.3 & 10.1 & 14.6 & 14.2 & 14.3 & 12.0 & 16.1 & 19.6 & 13.0 & 8.0 \\ \midrule
\textbf{Avg. Rel.~Passages / Q} & 2.0 & 2.1 & 2.6 & 2.8 & 4.3 & 2.1 & 2.0 & 2.1 & 2.1 & 3.1 & 2.2 & 2.6 & 2.8 & 1.9 & 1.2 & 1.8 & 1.2 & 2.5 \\
\textbf{Avg. Non Rel.~Passages / Q} & 8.1 & 8.0 & 7.7 & 7.7 & 5.6 & 8.3 & 8.1 & 7.9 & 7.8 & 7.0 & 8.2 & 11.8 & 7.6 & 8.7 & 4.9 & 8.5 & 8.8 & 7.5 \\ 

\midrule
\multicolumn{19}{c}{\textbf{\mirage Training Dataset}} \\ \midrule
\textbf{\#~Queries} & 3468 & 1624 & --- & 2857 & 2159 & 2104 & 2878 & 1137 & 1165 & 4054 & 3466 & 859 & 4567 & 1866 & 3283 & 2965 &  --- & 1311 \\
\textbf{\# Avg. Query Tokens} & 10.2 & 17.4 &  ---  & 8.7 & 11.2 & 14.2 & 11.8 & 10.4 & 17.1 & 9.9 & 14.5 & 14.9 & 14.3 & 12.0 & 16.1 & 19.3 &  --- & 7.9 \\ \midrule
\textbf{Avg. Rel.~Passages / Q} & 1.8 & 2.3 & --- & 2.5 & 3.6 & 2.0 & 1.7 & 2.0 & 2.0 & 2.9 & 1.9 & 2.1 & 2.0 & 1.4 & 1.2 & 1.6 & --- & 2.3 \\
\textbf{Avg. Non Rel.~Passages / Q} & 5.5 & 7.9 & --- & 7.5 & 5.3 & 8.3 & 5.3 & 8.0 & 7.9 & 7.1 & 7.9 & 12.4 & 5.2 & 3.5 & 4.3 & 5.6 & --- & 7.6 \\ \midrule
\textbf{\# Avg. GPT-4o Context Tokens} & 105.5 & 144.5 & --- & 137.9 & 203.4 & 133.8 & 129.4 & 180.1 & 149.4 & 100.8 & 140.2 & 136.4 & 90.0 & 106.4 & 54.6 & 86.0 & --- & 137.9 \\
\textbf{\# Avg. GPT-4o Answer Tokens} & 35.7 & 22.9 & --- & 20.6 & 55.3 & 51.6 & 30.3 & 56.0 & 26.6 & 22.6 & 24.1 & 23.9 & 16.1 & 18.4 & 11.4 & 20.4 & --- & 40.6 \\
\bottomrule
\end{tabular}}
\caption{Dataset statistics for 18 languages in \mirage; All tokens are calculated using the GPT-4o tokenizer \cite{openai:2023}; (Rel.) denotes relevancy; (\#~Avg GPT-4o) counts the tokens in the GPT-4o generated context and answer; Training data is not available for two languages (denoted by ---): German (de) and Yoruba (yo).}
\vspace*{-\baselineskip}
\label{table:mirage-statistics}
\end{table*}

We develop \mirage, a RAG benchmark across 18 languages for multilingual generation evaluation on Wikipedia. \mirage data is sourced from \miracl~\cite{zhang:2023}, a multilingual retrieval dataset containing human-generated queries and human-labeled relevance judgments for Wikipedia passages. We benchmark 19 frontier multilingual LLMs in our experiments. 
Our evaluation flowchart adopted in \mirage is shown in \autoref{fig:MIRAGE-flowchart}.
We evaluate seven heuristic features: (i) language detection, (ii) citation quality, (iii) support, (iv) reranker score, (v) answer overlap (traditional), (vi) answer overlap (LLM-measured), and (vii) fluency (LLM-measured). 
We use GPT-4o as the judge to evaluate our pairwise RAG comparisons on a smaller subset consisting of 100 queries.
Next, we train a random forest model as a surrogate judge, using the heuristic features as input, and learn to output the Bradley-Terry model logits as output. 
Finally, during inference, we use our surrogate judge to produce a ``synthetic'' ranking leaderboard for all baselines across every language. 

More specifically, we ask the following research questions in our work: 
\setlist{nolistsep}
\begin{itemize}[noitemsep]
    \item Can we avoid the LLM as a judge cost and combine heuristic-based evaluation? \vspace{1mm}
    \item How do frontier multilingual-focused LLMs perform in multilingual answer generation? \vspace{1mm}
    \item Does fine-tuning on \mirage training dataset improve LLM performance? \vspace{1mm}
\end{itemize}

\noindent Our experimental results show that: (i) random forest model (surrogate judge) learned rankings highly correlates with GPT-4o as a judge, achieving an average Kendall-Tau ($\tau$) score of 0.909. (ii) proprietary and large open-source LLMs ($\geq$70B parameters) achieve the topmost ranks in the \mirage leaderboard. (iii) \mirage training data, synthetically constructed using GPT-4o, is beneficial to improve smaller open-source LLMs (7--8B parameters). The main contribution of work is building \mirage and benchmark 19 frontier multilingual LLMs to accelerate development in the area of multilingual RAG.


\section{Related Work}

Prior work on RAG evaluation has been conducted exclusively in English. For example, benchmarks such as ALCE \cite{gao:2023}, FreshLLM \cite{vu:2023}, ClapNQ \cite{rosenthal:2024}, HAGRID \cite{kamalloo:2023} and CRAG \cite{yang:2024}, all include long-form answers for English-only queries and are based on collections containing documents from either the English Wikipedia, MS MARCO \cite{bajaj:2016} or NQ \cite{kwiatkowski:2019}. Similarly, TREC 2024 RAG, a TREC competition for RAG evaluation is focused on evaluating queries in English.\footnote{TREC 2024 RAG: \href{https://trec-rag.github.io/}{https://trec-rag.github.io/}}

\smallskip
\noindent\textbf{Multilingual RAG.} On the multilingual side, RAG has not been well studied in prior literature. The RGB benchmark~\cite{chen:2024} is limited in language scope covering only one additional language: Chinese (zh). NeuCLIR 2024 track \cite{mayfield:2024} evaluates long-form report generation from participants, but is limited to three languages. A concurrent work is BERGEN \cite{chirkova:2024}, which evaluates the multilingual open-domain QA setting on 13 languages. In contrast, in \mirage, we (i) evaluate the generation task
in the multilingual RAG pipeline on 18 languages, (ii) provide multilingual instruction-tuned data for RAG fine-tuning, and (iii) utilize high-quality native-speaker multilingual queries generated in \miracl \cite{zhang:2023}.

\smallskip
\noindent\textbf{Learning to rank.} A supervised learning technique, where models are trained to rank items within a list similar to training data ~\cite{liu:2010, learning-to-rank-101}. Models are trained in either a pointwise, pairwise, or listwise objective \cite{cao:2007}. In our work, we experiment with simple models such as random forest to train to rank the Bradley-Terry model coefficient produced by LLM as a judge. We keep complex models such as LambdaMART \cite{burges:2010} for future work.

\section{An Overview of \mirage}
We select 18 languages in \mirage as the starting point, representing an appropriate cross-section of the diversity of the languages spoken worldwide at this point. \mirage is a comprehensive multilingual RAG benchmark focusing on the generation task evaluation. 
As shown in \autoref{table:mirage-statistics}, the benchmark includes 11,195 evaluation pairs and 39,763 training pairs across 18 languages. 

\subsection{Distinction and Extension from \miracl} \miracl introduced in \citet{zhang:2023} is a monolingual retrieval dataset, which evaluates the retrieval task, i.e.,~given a user query and a passage corpus, retrieve the ranked list of passages as output. \miracl contains human-annotated relevance judgments to evaluate retrieval and re-ranker models, e.g., lexical models like BM25 \cite{bm25} or bi-encoders like mDPR \cite{karpukhin:2020}, or late-interaction models like ColBERT \cite{khattab:2020}. 

In contrast, \mirage evaluates the generation task in RAG, requiring LLMs to generate a summarized answer given the query and context available from oracle-judged passages. In our work, we reuse the queries and relevance-judgments from \miracl, but solely evaluate the multilingual generation task in RAG measuring answers using both heuristics and LLM as a judge. Since \mirage evaluates the generation task containing the oracle context, independent of the retrieval task, the chances of contamination in \mirage evaluation from fine-tuning \miracl is less overall.

\subsection{\mirage Evaluation Dataset} 
\miracl queries are high-quality and generated by native speakers \cite{zhang:2023}. The annotation procedure in \miracl is identical to TyDI-QA \cite{clark:2020}. The passage collection is constructed from language-specific Wikipedia corpora and parsed using WikiExtractor. 
The \mirage evaluation dataset is constructed re-using the queries and oracle-judged passages available in the \miracl development split.\footnote{We did not utilize the test split in MIRACL as the relevance judgments are not publicly available.} 

We incorporate two changes: (i) In Arabic (ar), we randomly sample a smaller subset of 1,501 out of 2,896 queries for uniformity in the number of queries available across other languages. (ii) We filter out a small subset of queries with zero non-relevant passages, i.e., we always include queries with hard passages from \miracl to make the \mirage evaluation challenging.\footnote{An exception is Telugu (te), where 78 queries have at least one non-relevant passage. Therefore, we randomly sample 72 additional queries with only relevant judged passages.}

\subsection{\mirage Training Dataset} 
The \mirage training dataset is developed from the \miracl training dataset \cite{zhang:2023}.
In \mirage, we reuse all the \miracl training pairs available in 16 languages (except German (de) and Yoruba (yo)) and convert it using a simple recipe into a multilingual instruction-tuned RAG dataset \cite{zhang:2024, niu:2024}. In our recipe, we first keep only the relevant passages as context along with the input query to generate a zero-shot RAG output by prompting strong teachers such as GPT-4o \cite{openai:2023}, Llama~3 (70B) \cite{llama3:2024} and Mixtral (8$\times$22B) \cite{jiang:2024}. After generation, we include non-relevant passages within our prompt as ``distracting and noisy context'', to help improve the quality of the training dataset. Since we convert a retrieval dataset, we do not have human-annotated answers for questions in \mirage.

\begin{figure*}[t]
\centering
\begin{center}
    \includegraphics[trim=0 0 0 0,clip,width=\textwidth]{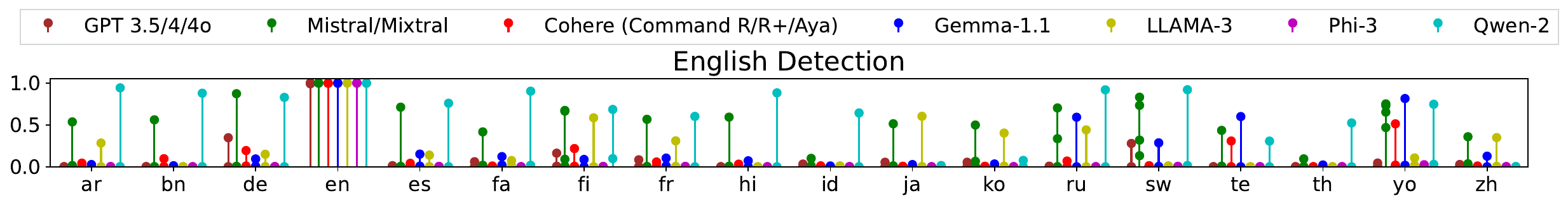}
    \includegraphics[trim=0 0 0 0,clip,width=\textwidth]{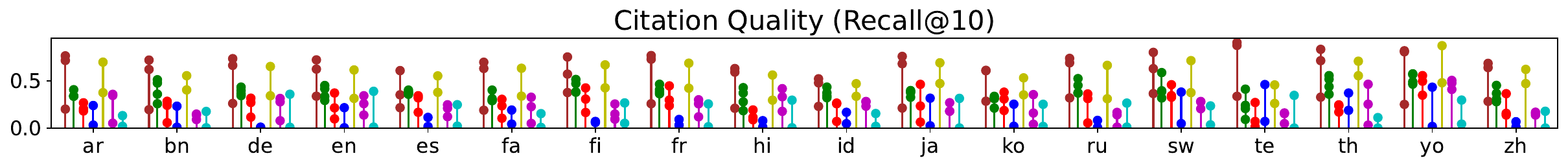} 
    \includegraphics[trim=0 0 0 0,clip,width=\textwidth]{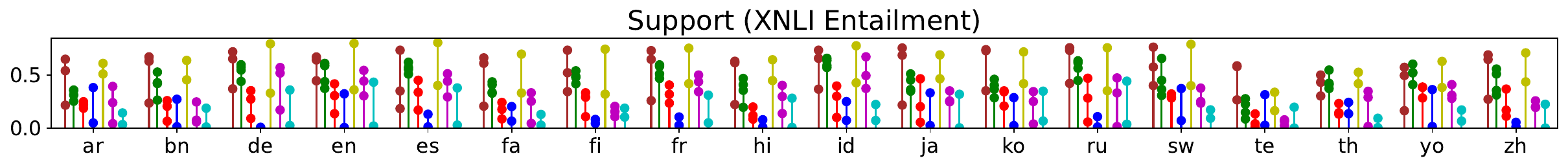}
    \includegraphics[trim=0 0 0 0,clip,width=\textwidth]{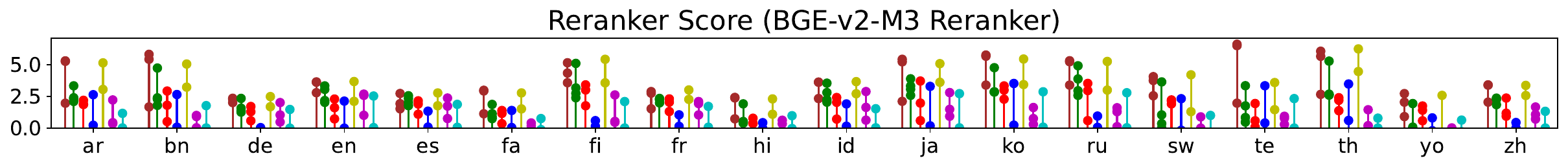}
    \includegraphics[trim=0 0 0 0,clip,width=\textwidth]{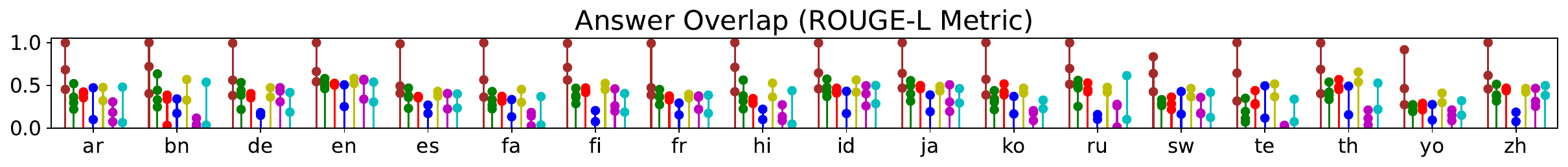}
    \includegraphics[trim=0 0 0 0,clip,width=\textwidth]{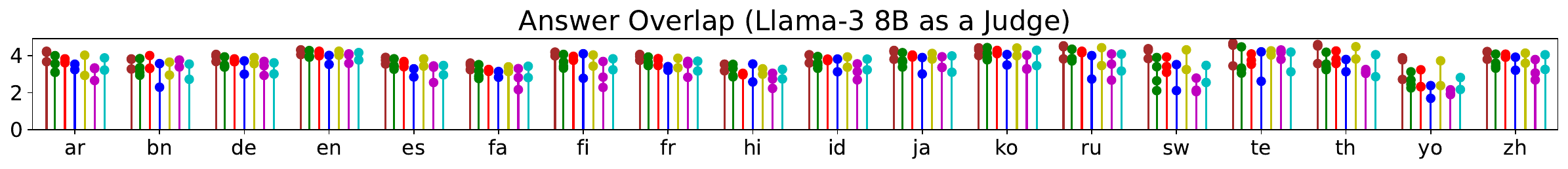}
    \includegraphics[trim=0 8 0 0,clip,width=\textwidth]{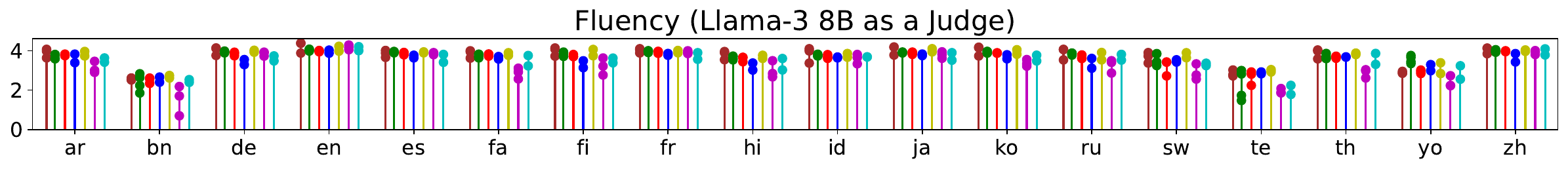}
    \caption{Lollipop plots denoting the average heuristic-based feature scores achieved by LLM baselines for each language in \mirage. $x$-axis denotes the 18 languages; whereas $y$-axis plots every heuristic feature score. Models in the same LLM family are represented in the same color in a lollipop (as multiple circles). \autoref{fig:heuristics-appendix} in the Appendix provides lollipop plots for all eleven heuristic-based features used in our work.}
    \label{fig:heuristics}
\end{center}
\vspace*{-\baselineskip}
\end{figure*}

\section{Multilingual RAG Evaluation}

\subsection{Heuristic-based Evaluation}
\label{sec:hbeval}
Answer generation in RAG requires evaluation on various dimensions. For example, whether a system's response provides the correct final answer or cites the relevant documents, a single metric alone is \emph{not sufficient} to capture the comprehensive evaluation required for RAG systems. Inspired by other recent works \cite{kiela:2021, santhanam:2023, gao:2023}, we introduce five \emph{deterministic} features and two \emph{LLM-measured} features for evaluation in our work. We rely on features that are explainable, cheap, and fast to compute. 
We explain each heuristic feature in \autoref{sec:heuristic-based-eval}.

\smallskip
\noindent\textbf{Language detection.} We compute the probability of a system's response in the required target language with \texttt{langid} \cite{lui:2012}. We compute two metrics: language detection (target language) and English detection.

\smallskip
\noindent\textbf{Citation quality.} Using passage-level relevance judgments for all queries (or qrels) information available in \miracl, we evaluate whether the system's response cites the relevant passages, crucial for measuring faithfulness. We compute and evaluate: Recall$@k$ and MAP$@k$, where $k$ = 10, as we have a maximum of 10 passages per query.

\smallskip
\noindent\textbf{Support.} Grounding is necessary to avoid hallucinations in the system's response. Support evaluation \cite{gao:2023} checks whether each sentence is supported by cited passages using a multilingual NLI model~\cite{he:2023}.\footnote{ \href{https://huggingface.co/MoritzLaurer/mDeBERTa-v3-base-xnli-multilingual-nli-2mil7}{MoritzLaurer/mDeBERTa-v3-base-xnli-multilingual-nli-2mil7}} We compute the probability of the \emph{entailment} and \emph{neutral} score, macro-averaged across the sentence-citation pairs.

\smallskip
\noindent\textbf{Reranker score.} The reranker score measures the average similarity (can be greater than 1.0) between the query and the passages cited within the system's response. We compute the reranker score using a multilingual reranker model,\footnote{Reranker \cite{chen:2024b}: \href{https://huggingface.co/BAAI/bge-reranker-v2-m3}{BAAI/bge-reranker-v2-m3}} macro-averaged across the query-passage pairs.

\smallskip
\noindent\textbf{Answer overlap.} Having the correct answer is crucial in the RAG system's response. Since \mirage does not include a human-labeled answer, we use the generated answer from GPT-4 \cite{openai:2023} as the gold truth. We compute two traditional metrics: SacreBLEU \cite{papineni:2002} and ROUGE-L \cite{lin:2004} measuring the lexical word overlap between the gold answer (GPT-4's answer is used for reference) and the system's response. 

\smallskip
\noindent\textbf{Answer overlap (LLM-measured).} In addition, we evaluate using Llama-3 (8B) \cite{llama3:2024}, an open-source LLM as a judge evaluator in a pointwise setup, providing a semantic word overlap integer score in the range $[1, 5]$. The answer overlap prompt description is listed in \autoref{fig:prompt_answer}.

\smallskip
\noindent\textbf{Fluency (LLM-measured).} It measures for grammatical correctness and idiomatic word choices in the system's response. As previously mentioned, we use Llama-3 (8B) \cite{llama3:2024} in a pointwise setup, proving an integer score in $[1, 5]$. 
The fluency prompt description is listed in \autoref{fig:prompt_fluency}.

\subsection{Arena-based Evaluation}
Heuristic-based evaluation metrics often rely on a gold standard for evaluation. Tasks such as text retrieval \cite{bajaj:2016, thakur:2021} require human-labeled relevance judgments, and similarly, NLP tasks such as machine translation \cite{stahlberg:2020}, require human-annotated translations. As human preferences are seldom available in numerous applications, using LLM as a judge \cite{zheng:2024, chen:2024c, chiang:2024} is becoming a de facto approach for arena-based evaluation of LLMs. 

We evaluate pointwise, listwise, and pairwise LLM as judge evaluations. We anecdotally observe the pointwise judge, which is efficient $O(n)$, but is not good for ranking LLMs as it provides similar scores for a wide range of models (e.g., 4/5 score for 16 out of 18 LLMs evaluated). Similarly, the listwise judge finds it difficult to rank all 19 models in the correct order. Therefore, although suboptimal in complexity, $O(n^2)$, where $n$ is the number of models evaluated, we choose a pairwise evaluation in our work.

\smallskip
\noindent\textbf{Pairwise LLM as a judge.} Following prior works on arena-based evaluation in RAG \cite{rackauckas:2024, pradeep:2024}, we evaluate two system's responses in a head-on comparison by computing pairwise judgments with LLM as a judge.
We reuse the RAGElo prompt template \cite{rackauckas:2024} with minor additions.  The prompt template is listed in \autoref{fig:prompt_gpt4}. LLM as a judge evaluator includes three types of biases \cite{zheng:2024}: (i) verbosity bias \cite{wu:2023} (ii) self-enhancement bias \cite{xu:2024, panickssery:2024} and (iii) position bias \cite{wang:2023}. 
We avoid the verbosity bias, as RAG evaluation has fixed evaluation criteria requiring sentence-level citations and answers \cite{pradeep:2024} and the position bias by randomly swapping the position of two models. 

\subsection{Learning to Approximate Rankings}
\label{sec:learn_to_approx}
There is no predefined way to aggregate the heuristic features to provide an overall leaderboard ranking in \mirage. Averaging the scores is too simplistic as features measure different aspects of RAG evaluation. On the other hand, arena-based evaluations provide ranked leaderboards but are computationally expensive to compute with a strong teacher model. To avoid computational costs, smaller models as teachers have been proposed \cite{thakur:2024b, ni:2024}. 

Motivated by similar observations, we train a surrogate judge to effectively emulate an arena-based leaderboard without incurring the expensive LLM as a judge pairwise cost. We find the random forest model to serve as a scalable and cost-effective judge that can be trained within minutes on small training datasets without expensive computation. Therefore, we train a random forest (learning to rank model) as a surrogate judge to approximate the Bradley-Terry model coefficients \cite{hunter:2004} learned from an arena-based evaluation that uses GPT-4o as a judge for pairwise judgments. 

\begin{figure*}[t]
\centering
\begin{center}
    \includegraphics[trim=0 0 0 0,clip,width=0.48\textwidth]{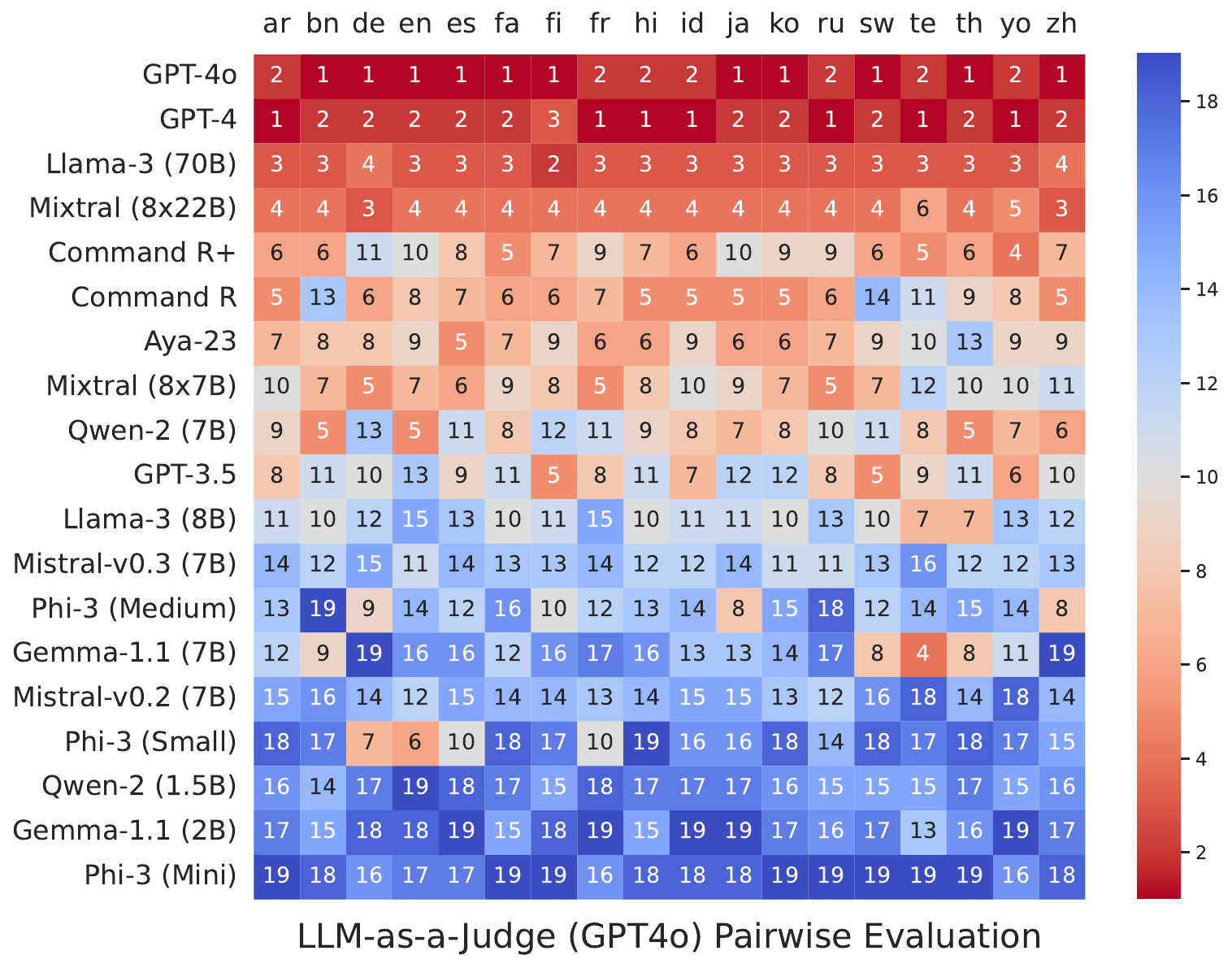}
    \includegraphics[trim=0 0 0 0,clip,width=0.48\textwidth]{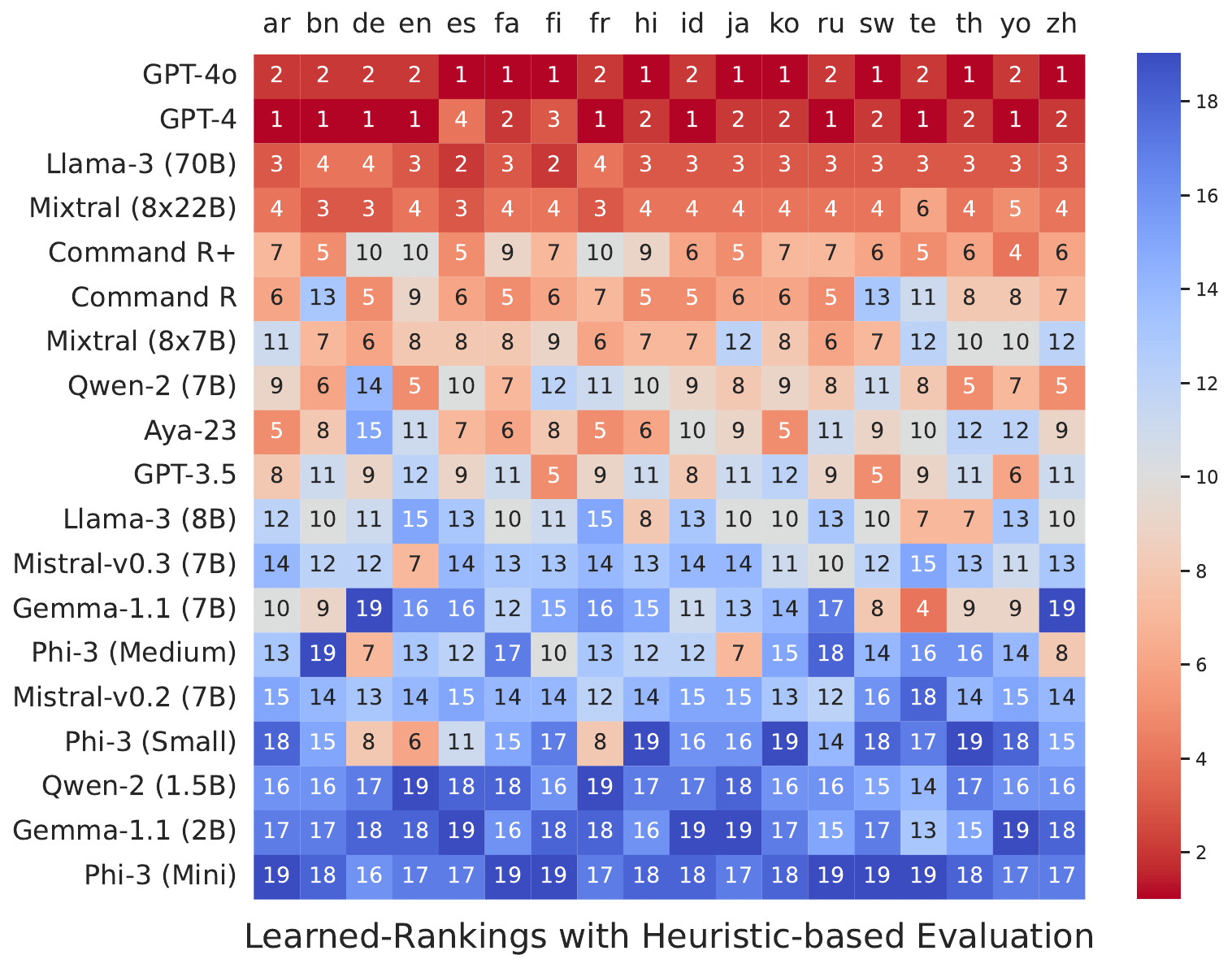}
    \caption{\mirage arena-based leaderboards: (left heatmap) Bradley-Terry model coefficients with GPT-4o as a pairwise judge for a subset of 100 sampled queries; (right heatmap) Synthetic rankings using heuristic-based features and a random forest model as a surrogate judge on all queries. Each highlighted cell denotes the rank of the LLM (lower the better). LLMs are sorted by lowest to highest average rank across all 18 languages.}
    \label{fig:arena-based-rankings}
\end{center}
\vspace*{-\baselineskip}
\end{figure*}

\smallskip
\noindent\textbf{Learning to rank model.}
While the heuristic features introduced in Section \ref{sec:hbeval} can be computed efficiently and without the reliance on proprietary LLMs, inducing a ranking from pairwise comparisons via a Bradley-Terry model is computationally expensive and requires access to a high-performance LLM. Furthermore, as we demonstrate in Section \ref{sec:ablations}, the ranking accuracy, measured by the Kendall-Tau ($\tau$) coefficient, degrades rapidly when subsampling tournament matches. 

The procedure, detailed in Algorithm \ref{alg:simulate_tourney}, simulates $N_t$ tournaments, each involving a total of $N_l$~models and $N_q$ queries. For each query, judgments are obtained for all $\binom{N_l}{2}$ pairings of models. We employ bootstrapping on the query selection process to estimate the variance in the $R^2$ metric in the learning to rank models' approximations of the Bradley-Terry coefficients over a randomly-sampled holdout set, $\textsc{llm}_{predict}$. 

We randomly select two models, Gemma 1.1 (2B) and Llama-3 (70B) as holdout models, i.e., we do not train on the features for holdout models. For English, we observe an average $\bar{R}^2 = 0.971$ with a 95\% confidence interval of $[0.905, 0.999]$. On the other hand, for Bengali, we observe $\bar{R}^2 = 0.937$ with a 95\% confidence interval of $[0.766, 0.998]$. $\bar{R}^2$ scores for all 18 languages with 95\% confidence intervals are listed in \autoref{table:r2_scores}. Taken together, these results indicate that the training procedure is fairly robust with $N_q=100$.

\begin{algorithm}[tbh]
\caption{Simulate Tournaments and Fit Models}
\begin{algorithmic}[1]
\label{alg:simulate_tourney}

\FOR{$i \in [N_t]$}

    \STATE $\mathit{M_{BT}^i} \gets \textsc{Tournament}(N_q)$

    \STATE $X_{t}, Y_{t} \gets \textsc{Dataset}(\textsc{llm}_{train}, M_{BT}^i)$
    
    \STATE $X_{p}, Y_{p} \gets \textsc{Dataset}(\textsc{llm}_{predict}, M_{BT}^i)$
    
    \STATE $\mathit{M_{reg}^i} \gets {\textsc{Fit}(X_{t}, Y_{t})}$
    
    \STATE {$R_{2}^i \gets \mathit{M_{reg}^i}(X_{p}, Y_{p})$}
    
\ENDFOR
\STATE $\mathit{M_{BT}} \gets [M_{BT}^1;M_{BT}^2;...;M_{BT}^{N_t}]$
\STATE $\mathit{M_{reg}} \gets [M_{reg}^1;M_{reg}^2;...;M_{reg}^{N_t}]$
\STATE $R_2 \gets [R_2^1; R_2^2; ...; R_2^{N_t}]$
\end{algorithmic}
\vspace{1em}  
\footnotesize  
\textbf{Note:} Refer to Section \ref{sec:learn_to_approx} for a definition of each of the variables. The \textsc{Tournament} function runs a battle arena, sampling $q$ queries, and returning the learned Bradley-Terry model. The \textsc{Dataset} function accepts a set of LLMs and a learned Bradley-Terry model. It returns $X$, the heuristic RAG feature values, and $Y$, the Bradley-Terry coefficients for each LLM model. After simulating $N_t$ tournaments, the array $R_2$ contains the $R^2$ errors for each of the $N_t$ models.
\end{algorithm}

\section{Experimental Settings}
\subsection{Multilingual Baselines}
Existing frontier LLMs are either English-only or support a limited set of languages, predominantly due to the \emph{curse of multilinguality} for large models \cite{conneau:2020}. It is unclear how well existing LLMs perform on RAG on a wide variety of languages, due to the scarce availability of multilingual instruction tuning datasets. We experiment with LLMs from seven different families, containing proprietary and open-source LLMs. Wherever possible, we evaluate the \emph{instruction-tuned} version if available. Refer to \autoref{sec:additional-details} for more details.

\begin{itemize}[leftmargin=*, topsep=2pt, ]
    \setlength\itemsep{0.0em}
    \item \textbf{OpenAI:} GPT-3.5-turbo, GPT-4, and GPT-4o \cite{openai:2023} using the Azure OpenAI service.
    \item \textbf{Mistral:} Mistral-Instruct-v0.2 and v0.3 (7B)~\cite{jiang:2023} and Mixtral-Instruct-v0.1 (8$\times$7B) and (8$\times$22B) versions~\cite{jiang:2024}.
    \item \textbf{Cohere:} Command-R (35B), Command-R+ (104B) and Aya-23 (35B)~\cite{aryabumi:2024}.
    \item \textbf{Gemma:} Gemma 1.1 instruct (2B) and (7B) models \cite{mesnard:2024}.
    \item \textbf{Llama-3:} Llama-3 instruct (8B) and (70B) models \cite{llama3:2024}. 
    \item \textbf{Phi-3:} Phi 3 instruct series: Medium (14B), Small (7B), and Mini (3.8B) \cite{phi3:2024}.
    \item \textbf{Qwen-2:} Qwen-2-instruct series: 1.5B and 7B \cite{qwen2:2024}.
\end{itemize} 


\smallskip
\noindent\textbf{Prompt template.} We internally optimized\footnote{A majority of the prompt optimization was internal and based on eye-balling RAG responses across LLMs.} the ChatQA prompt template \cite{liu:2024}, to include in-text citations of the context passages following the IEEE format \cite{kamalloo:2023}. In \mirage, we have about $10$ passages annotated in the oracle setting. Therefore, we trim each passage available and take the first 800 tokens to fit all passages within a fixed context length of 8192 tokens. following prior work in \citet{shi:2023}, the prompt requires the LLM to explain the multilingual generation task starting with ``\texttt{\#\#Reason}'' and the answer itself starting with ``\texttt{\#\#Answer}''. Utilizing this output format has its advantages in easily parsing the generated answer and the rationale behind the answer. The prompt template for multilingual generation is shown in \autoref{fig:prompt_baseline}. 


\section{Experimental Results}

\begin{table}[t]
\centering
\resizebox{0.48\textwidth}{!}{
\begin{tabular}{cc|cc|cc|cc}
\toprule
\textbf{Lang.} & \textbf{$\tau$} & \textbf{Lang. } &  \textbf{$\tau$}  & \textbf{Lang. } &  \textbf{$\tau$} & \textbf{Lang. } &  \textbf{$\tau$} \\
\midrule
ar & 0.951 & bn & 0.874 & de & 0.825 & en & 0.835 \\
es & 0.876 & fa & 0.924 & fi & 0.949 & fr & 0.914 \\
hi & 0.946 & id & 0.896 & ja & 0.892 & ko & 0.950 \\
ru & 0.849 & sw & 0.958 & te & 0.938 & th & 0.946 \\
& & yo & 0.906 & zh & 0.941 & & \\ \midrule
\multicolumn{8}{c}{Avg. Kendall Tau ($\tau$) on 18 languages = \textbf{0.909}} \\
\bottomrule
\end{tabular}}
\caption{Kendall $\tau$ rank correlation between pairwise GPT-4o as a judge Bradley-Terry model and the synthetic arena-based ranking leaderboard generated using our surrogate judge in \mirage.}
\label{tab:kendall_tau}
\end{table}

\subsection{Heuristic-based Results}\label{sec:heuristic-results} 

\autoref{fig:heuristics} shows lollipop plots indicating the average heuristic-feature value ($y$-axis) distribution for each language ($x$-axis). In English detection (higher the worse), smaller LLMs such as Gemma-1.1 (2B) do not generate output in the required target language but rather generate reasoning and answers in English. For citation quality, support, and reranker score features, LLMs from OpenAI and Llama-3 family achieve high Recall$@10$ and entailment scores (except Llama-3 (70B) for a few languages), indicating generated answers include grounded citations from relevant passages. In contrast, LLMs from the Qwen-2 or Gemma-1.1 family, tend to under-cite passages in their answers.

Furthermore, we observe LLMs from the OpenAI family achieve the highest word overlap in the ROUGE-L metric (GPT-4 used as ground truth, which could be a potential bias) and Llama-3 (8B) as a judge, while we observe less variance across other LLMs. In Fluency, we observe a majority of the LLMs are rather fluent in generation, except Bengali (bn), Telugu (te), and Yoruba (yo).

\begin{figure}[t]
\centering
\begin{center}
    \includegraphics[trim=0 0 0 0,clip,width=0.49\textwidth]{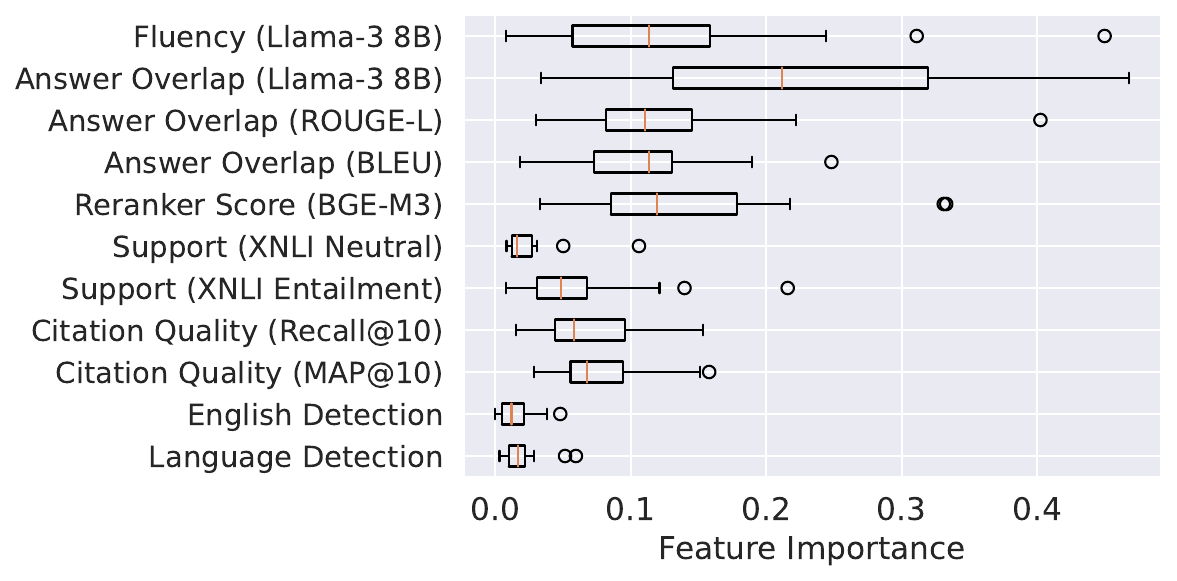}
    \caption{Boxplot with the feature importance value (averaged across 18 languages in \mirage) observed by the learning to rank (random forest) model.}
    \label{fig:feature-importance}
\end{center}
\vspace*{-\baselineskip}
\end{figure}

\subsection{Arena-based Results}\label{sec:arena-results} 
\autoref{fig:arena-based-rankings} (left heatmap) is the arena-based leaderboard with the Bradley Terry model conducting 200 tournaments and bootstrapping 100 matches per tournament on a subset of 100 queries using GPT-4o as a pairwise judge. We observe that proprietary LLMs such as GPT-4o and GPT-4, and larger open-source LLMs such as Llama-3 (70B) and Mixtral (8$\times$22B) perform better than other LLMs. LLM rankings across languages are usually stable; with a few notable exceptions such as Gemma-1.1 (7B) which achieves a rank of 4 in Telugu. Command R (35B) performs poorly in low-resource languages such as Bengali (rank 13) or Swahili (rank 14). The complete scores including model coefficient logits and 95\% confidence intervals (error bars) are provided in \autoref{table:bradley_terry_page1} and \autoref{table:bradley_terry_page2} in the Appendix.

\smallskip
\noindent\textbf{Synthetic rankings using random forest.} \autoref{fig:arena-based-rankings} (right heatmap) is the learned synthetic leaderboard rankings on all queries using heuristic-based features trained with a random forest model. The synthetic leaderboard generated using a surrogate judge, highly correlates to the GPT-4o as a pairwise judge, achieving an average Kendall-Tau ($\tau$) rank correlation = \textbf{0.909}, by training on 17 LLMs during training and keeping 2 LLMs as a holdout for every language.
Individual language-specific Kendall-Tau rank correlation scores are listed in \autoref{tab:kendall_tau}.
This provides evidence of the efficacy of training a random forest model as a surrogate judge.
In \autoref{sec:extending-eval}, we extend the evaluation for Llama-3.1 and Gemma-2 series of LLMs.

\smallskip
\noindent\textbf{Heuristic feature importance.} In \autoref{fig:feature-importance}, we plot the average feature importance achieved by our random forest model as a surrogate judge. Using Llama-3 (8B) as a judge, for fluency and answer overlap are the important heuristic features. Similarly, deterministic answer overlap and reranking-based metrics are equally important. Some heuristic features such as language detection, and neutral score in support evaluation obtain the least importance. We observe all answer-related heuristic features achieve a high importance indicating that the generated ``answer'' portion in an LLM's response is crucial and required to learn rankings from GPT-4o as a pairwise judge.

\begin{table}[t]
\centering
\resizebox{0.48\textwidth}{!}{
\begin{tabular}{lrrrrrrr}
\toprule
Model / Language               & ar    & bn    & fi    & ja    & ko    & ru    & te   \\
\midrule
\multicolumn{8}{c}{Train $R^2$ on randomly selected fifteen models} \\ \midrule
Random Forest & 0.97  & 0.96  & 0.97  & 0.96  & 0.97  & 0.95  & 0.97 \\
Linear Regression      & 0.98  & 0.98  & 0.98  &\textbf{ 1.00}  & 0.97  & 0.99  & \textbf{1.00} \\
MLP Regressor          & 0.97  & 0.98  & 0.97  & 0.96  & 0.96  & 0.98  & 0.99 \\
XGB Regressor         & \textbf{1.00}  & \textbf{1.00}  & \textbf{1.00}  & \textbf{1.00}  & \textbf{1.00}  & \textbf{1.00}  & \textbf{1.00} \\
SVR                   & 0.75  & 0.77  & 0.81  & 0.59  & 0.67  & 0.73  & 0.39 \\ \midrule
\multicolumn{8}{c}{Holdout $R^2$ on four randomly selected held out models} \\ \midrule
Random Forest          & \textbf{0.50}  & 0.41  & 0.49  & \textbf{-0.03} & \textbf{0.45}  & 0.07  & \textbf{0.83} \\
Linear Regression     & -1.92 & -5.45 & -19.38 & -0.06 & -26.53 & -2.13 & 0.31 \\
MLP Regressor                 & 0.33  & 0.37  & 0.45  & -0.76 & -0.04 & -0.48 & 0.78 \\
XGB Regressor               & -0.02 & 0.44  & 0.22  & -1.33 & -0.09 & -0.80 & 0.59 \\
SVR                   & -0.03 & \textbf{0.48} & \textbf{0.64} & -0.53 & -0.09 & \textbf{0.15}  & -0.10 \\
\bottomrule
\end{tabular}}
\caption{Train and Holdout $R^2$ scores using different learning to rank model choices. Each experiment has been repeated 50 times with four held-out models.}
\label{table:regression_model_ablation}
\end{table}

\subsection{Ablations \& Discussion}
\label{sec:ablations}

To better understand the gaps observed during training of the random forest model as a surrogate judge, we conduct further ablations on a subset of seven languages including Arabic (ar), Bengali (bn), Finnish (fi), Japanese (ja), Korean (ko), Russian (ru) and Telugu (te):

\smallskip
\noindent\textbf{Learning to rank model choice.} We compare different learning to rank models as choices for learning the Bradley-Terry model coefficients. We conduct our experiments on the train set, where models contain pairwise judgments, and on a randomly sampled holdout set, a realistic scenario, with no available training data. We evaluate the following choices: Random Forest, Linear Regression, MLP Regressor, XGB Regressor, and SVR. All models are implemented via \texttt{scikit-learn}.\footnote{\href{https://scikit-learn.org/stable/supervised\_learning.html}{https://scikit-learn.org/stable/supervised\_learning.html}} The results are shown in \autoref{table:regression_model_ablation}. Random forest achieves the best $R^2$ metric on the holdout subset for 4 out of 7 languages. SVR also achieves a similar $R^2$ metric on the holdout subset, however, underperforms random forest on the training subset. Other baselines, such as XGB Regressor and MLP Regressor show signs of significant overfitting on the training subset, thereby underperforming random forest on the holdout subset.

\begin{figure}[t]
\centering
\begin{center}
    \includegraphics[trim=0 0 0 0,clip,width=0.24\textwidth]{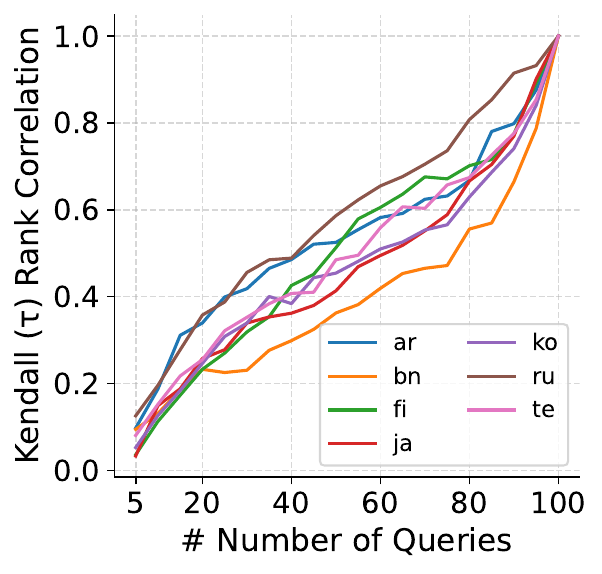}
    \includegraphics[trim=20 0 0 0,clip,width=0.22\textwidth]{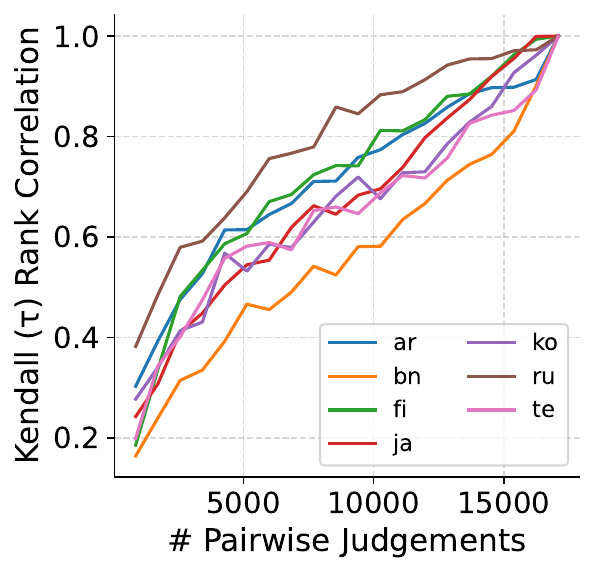}
    \caption{Sampling experiments to reduce computation cost. (left) reduces the number of queries whereas (right) reduces the pairwise judgments.}
    \label{fig:sampling}
\end{center}
\vspace*{-\baselineskip}
\end{figure}

\begin{table}[t]
\centering
\resizebox{0.48\textwidth}{!}{
\begin{tabular}{lcrrrrrrr}
\toprule
Features / Language & \#F      & ar     & bn     & fi    & ja      & ko     & ru    & te   \\ \midrule
(All)  Features         & 11   & 0.938  & \textbf{0.867}  & 0.921  & 0.881  & 0.921  & 0.826 & 0.918 \\
(W/o)  LLM as a Judge   &  9   & 0.912  & 0.866  & 0.898  & 0.853  & 0.891  & 0.811 & 0.904 \\
(W/o)  Low. Correlation &  7   & \textbf{0.951}  & \textbf{0.867}  & \textbf{0.923} & \textbf{0.885}  & \textbf{0.929}  & 0.829 & \textbf{0.940} \\
(Only) LLM as a Judge   &  2   & 0.948  & 0.728  & 0.907  & 0.884  & 0.916  & \textbf{0.851} & 0.872 \\
\bottomrule
\end{tabular}}
\caption{Kendall Tau ($\tau$) scores using different features for training the random forest regression model. }
\label{table:feature-based-ablation}
\end{table}

\smallskip
\noindent\textbf{Non-exhaustive pairwise comparisons lead to performance degradation.} Exhaustive pairwise comparisons across a subset of 19 models in \mirage using GPT-4o for all queries are quite expensive. To avoid this, we investigate whether all pairwise exhaustive comparisons are necessary during training. We utilize two sampling techniques: (i) full pairwise judgments on a subsample of 100 queries, e.g., 20 or 50 queries; (ii) partially judge a non-exhaustive random sample of the pairwise judgments across 100 queries, e.g., only 50\% of all the exhaustive pairwise combinations. Both results are shown in \autoref{fig:sampling}. we observe that Kendall-Tau ($\tau$) correlations increase linearly with queries and pairwise judgments. In summary, an exhaustive pairwise comparison and a sufficient number of queries, such as 100, are necessary without impacting the leaderboard rankings.

\smallskip
\noindent\textbf{All heuristic features are not necessary.} We experiment with features used for random forest model training as a surrogate judge. We evaluate four training configurations: (i) all features (ii) without LLM-measured features (iii) without language detection and support, i.e., the low-correlation features observed in \autoref{fig:feature-importance}, and (iv) including only LLM-measured features. From \autoref{table:feature-based-ablation}, we observe that removing low-correlated heuristic RAG features helps train the random forest model better leading to a conclusion that not necessarily all heuristic features are important. Removing the LLM-measured features completely or only using them for training the model decreases the Kendall-Tau ($\tau$) correlation score.

\begin{figure}[t]
\centering
\begin{center}
    \includegraphics[trim=0 0 0 0,clip,width=0.48\textwidth]{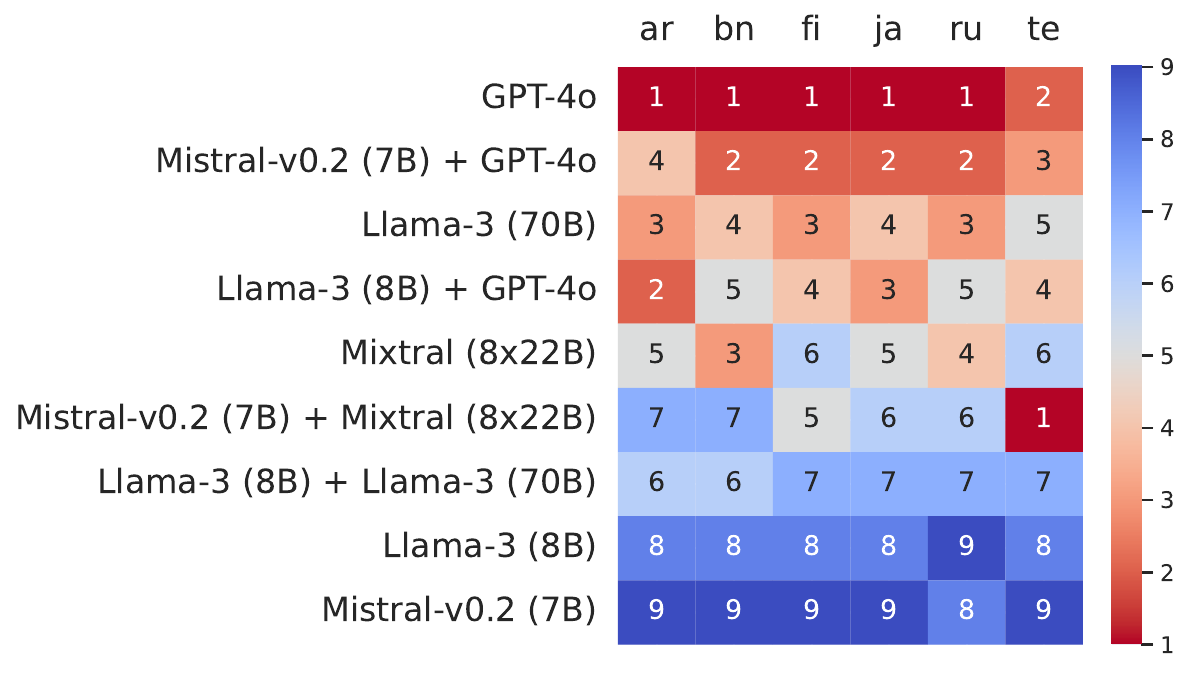}
    \caption{Approximate rankings using heuristic features after fine-tuning Llama-3 (8B) and Mistral-v0.2 (7B) on \mirage dataset across four configurations.}
    \label{fig:fine-tuning}
\end{center}
\vspace*{-\baselineskip}
\end{figure}

\smallskip
\noindent\textbf{Fine-tuning on \mirage training data.} 
We evaluate three variants of the \mirage training dataset using two smaller open-source LLMs: Mistral-v0.2 (7B) and Llama-3 (8B). We fine-tune the \mirage training datasets using (i) both on GPT-4o, (ii) Llama-3 (8B) on Llama-3 (70B), and (iii) Mistral-v0.2 (7B) on Mixtral (8$\times$22B). From \autoref{fig:fine-tuning}, we observe that GPT-4o is a strong teacher, Mistral-v0.2 (7B) fine-tuned on GPT-4o distilled training data achieves rank 2 outperforming Llama-3 (70B). This shows that \mirage training data is useful for improving the RAG answer generation task quality.

\section{Conclusion}
We present \mirage, a multilingual RAG benchmark for 18 languages aimed at evaluating the multilingual generation part within RAG and aggregate traditional heuristic-based features to train a lightweight learning to rank model as a surrogate judge to learn a Bradley Terry model with GPT-4o pairwise judgments. Our results indicate a strong correlation between our surrogate judge and GPT-4o as a pairwise judge. This demonstrated the effectiveness of our efficient, cheap, and easy-to-retrain random forest model as a surrogate judge trained using only computationally cheap heuristic features for arena-based leaderboard ranking by achieving a 0.909 Kendall $\tau$. 
On \mirage, we observe that most proprietary and open-source larger LLMs currently dominate, whereas smaller open-source LLMs continue to struggle. Instruction tuning on \mirage training data helps improve smaller open-source LLMs, e.g. instruction-tuned Mistral-v0.2 (7B) on GPT-4o distilled training data can outperform Llama 3 (70B) on \mirage. 

\section{Limitations}
\mirage is one of the first large-scale multilingual RAG benchmarks. Although not perfect, we below discuss a set of limitations in our work:
\begin{itemize}[leftmargin=*, topsep=2pt, ]
    \setlength\itemsep{0.0em}
    \item In \mirage, we focused on benchmarking the generation task in RAG with oracle passages, we did not consider the retrieval task and its error propagating on the generation task. 
    \item Due to budget constraints, we were unable to evaluate diverse LLMs as teachers such as Claude-3.5 (sonnet) \cite{claude-35-sonnet} or Gemini Pro \cite{anil:2023}. We evaluated using GPT-4o which can cause self-enhancement bias towards LLMs in the OpenAI family.
    \item In our heuristic evaluation, we only considered a smaller subset of features. We did not explore more recent hand-crafted features such as nugget-based recall and precision \cite{pradeep:2024b, farzi:2024, arabzadeh:2024, lin:2005}.
    \item Lastly, \mirage does not provide human-labeled answers for queries across all languages and is limited to Wikipedia as the source.
\end{itemize}

\section*{Acknowledgements}
This research is supported by the Canada First Research Excellence Fund and the Natural Sciences and Engineering Research Council (NSERC) of Canada. We thank Ka Wong for helping us out with bootstrapping during the training of the random forest model as a surrogate judge.

\bibliography{custom}
\bibliographystyle{acl_natbib}

\clearpage

\appendix

\section{Heuristic-based Evaluation: Features \& Additional Details}\label{sec:heuristic-based-eval}

\smallskip
\noindent\textbf{1. Language Identification:} In a multilingual RAG system, the output response should ideally be in the same language in which the user asked their query. To capture this feature, we attempt to identify which natural language the output response is in. We use \texttt{langid} \cite{lui:2012}, an off-the-shelf language detecting Python library for detecting the language of the long-form RAG answer. We use the probability of the target language detected as the score for language identification, i.e.,~$\hat{p} = langid(a, t)$, where $t$ denotes the target language and $a$ denotes the long-form answer.

\smallskip
\noindent\textbf{2. Citation Quality:} A multilingual RAG system must cite information from the relevant passages within their answers, to improve faithfulness and reduce hallucinations. We capture whether the passages (using relevance judgments provided in \miracl \cite{zhang:2023}) are cited in the multilingual generation task. For scoring, we compute the Recall$@k$ and MAP$@k$ score, where Recall$@k$ is 1.0 for a generated answer $a$, if and only if $a$ cites all available relevant passages, 
Similarly, the MAP$@10$ score measures the percentage of relevant passages within the top-$k$ cited passages.

\smallskip
\noindent\textbf{3. Support:} RAG systems have been shown to hallucinate across retrieval-augmented generation tasks, especially when provided with non-relevant contexts \cite{thakur:2024c}. Grounding is necessary to avoid hallucinations. We compute the grounding score of every sentence $s_{j}$ in generated answer $A$ along with the cited context $c_{j}$ using the multilingual NLI model, which computes the similarity score as a probability of either \emph{entailment}, \emph{neutral} or \emph{contradiction}. The entailment denotes the generated sentence in the long-form answer, which entails the cited passage within its response.

\smallskip
\noindent\textbf{4. Reranker Score:} The reranker score measures the semantic similarity between the user query and the cited passages in the system's response. If the cited passages are relevant in answering the query, the reranker model would output a higher similarity score. We utilize a multilingual open-source reranker, namely BGE-M3 for our evaluation. We compute the reranker score across each cited passage $p^j_{i}$ included in the long-form answer along with the user query $q_i$.

\smallskip
\noindent\textbf{5. Answer Overlap:} Existing open-domain question answering datasets such as Natural Questions \cite{kwiatkowski:2019}, HotpotQA \cite{yang:2018} or ELI5 \cite{fan:2019} all include a human-labeled answer, assisting in evaluation using text overlap metrics such as Exact Match (EM) or F1. However, user queries in RAG systems potentially generate long-form answers, requiring metrics such as SacreBLEU or ROUGE-L to evaluate text-generation tasks. Automatic metrics are fairly quick and cheap to compute. For this reason, we include two metrics, SacreBLEU and ROUGE-L, for evaluating the RAG-generated answer. As we do not have human-labeled answers in \mirage, we consider the GPT-4 generated answer as the gold truth for evaluation. 

\smallskip
\noindent\textbf{6. Answer overlap (LLM-measured):} To capture semantic overlap between answers, we use the Llama-3 (8B) model as the judge for evaluation in a pointwise setup, where the LLM as a judge outputs a score between 1 to 5.

\noindent\textbf{7. Fluency (LLM-measured):} Fluency measures for grammatical correctness and idiomatic word choices in long-form answer generation. Evaluating fluency in multilingual long-form generation answers is not straightforward. While existing techniques are available for English such as MAUVE \cite{pillutla:2021}, only a few models evaluate multilingual summarization \cite{clark:2023}. Inspired by recent works in G-EVAL \cite{liu:2023}, we evaluate fluency using open-source LLM such as Llama-3 (8B) as the judge. Our reason for choosing open-source models lies in reducing the expense, of running an expensive proprietary LLM such as GPT-4. Our LLM as a judge setup outputs a score between 1 to 5. 

\section{Baselines: Additional Details}\label{sec:additional-details}
In this section, we briefly describe each of the 19 multilingual-focused models utilized in our \mirage evaluation experiments:

\begin{enumerate}[leftmargin=*, topsep=2pt, ]
    \setlength\itemsep{0.0em}
    \item \textbf{GPT-3.5-turbo:} \cite{openai:2023} is evaluated using the Azure OpenAI service.\footnote{\href{https://learn.microsoft.com/en-us/azure/ai-services/openai/}{https://learn.microsoft.com/en-us/azure/ai-services/openai/}} We set the temperature parameter to $0.1$ for a deterministic output. It utilizes the \texttt{cl100k\_base} BPE-based tokenizer in the tiktoken\footnote{\href{https://github.com/openai/tiktoken}{https://github.com/openai/tiktoken}} repository.
    \item \textbf{GPT-4:} \cite{openai:2023} is also evaluated using the Azure OpenAI service. We use a temperature setting of $0.1$ for a deterministic output and the \texttt{cl100k\_base} BPE-based tokenizer.
    \item \textbf{GPT-4o:} \cite{openai:2023} is also evaluated using the Azure OpenAI service. We use a temperature setting of $0.1$ for a deterministic output and the \texttt{o200k\_base} BPE-based tokenizer.
    \item \textbf{Mistral-7B-Instruct-v0.2} \cite{jiang:2023} is the v0.2 of the instruct-version model containing 7B parameters.\footnote{\href{https://huggingface.co/mistralai/Mistral-7B-Instruct-v0.2}{mistralai/Mistral-7B-Instruct-v0.2}}~It is an English-centric model, i.e., not instruction fine-tuned with any multilingual data. We used the multiple GPU inference using the \texttt{vllm} repository \cite{kwon:2023}. We set the temperature parameter to 0.1.
    \item \textbf{Mistral-7B-Instruct-v0.3} \cite{jiang:2023} is an extension of the instruct-version 0.2 model containing 7B parameters.\footnote{\href{https://huggingface.co/mistralai/Mistral-7B-Instruct-v0.3}{mistralai/Mistral-7B-Instruct-v0.3}}~We set inference parameters similar to the previous model.
    \item \textbf{Mixtral-8$\times$7B-Instruct-v0.1} \cite{jiang:2024} is a pretrained generative sparse Mixture of Experts (MoE), containing 8$\times$7B parameters. It has been pretrained in 5 languages including English, French, Italian, German, and Spanish.\footnote{\href{https://huggingface.co/mistralai/Mixtral-8x7B-Instruct-v0.1}{mistralai/Mixtral-8x7B-Instruct-v0.1}}~As the model is computationally not feasible to evaluate due to resource constraints, We use the model API endpoint available in the Anyscale platform (\href{https://www.anyscale.com/}{https://www.anyscale.com/}), with a temperature setting of 0.1.
    \item \textbf{Mixtral-8$\times$22B-Instruct-v0.1} \cite{jiang:2024} is a pretrained generative sparse Mixture of Experts (MoE), containing 8$\times$22B parameters. Similar to before, it has pretrained on 5 languages including English, French, Italian, German, and Spanish.\footnote{\href{https://huggingface.co/mistralai/Mixtral-8x22B-Instruct-v0.1}{mistralai/Mixtral-8x22B-Instruct-v0.1}}~We utilize the model API endpoint available in the Anyscale platform (\href{https://www.anyscale.com/}{https://www.anyscale.com/}), with a temperature setting of 0.1.
    \item \textbf{Command R} is developed keeping RAG in mind and officially supports 11 languages: Arabic, Brazilian, Portuguese, English, French, German, Italian, Japanese, Korean, Chinese, and Spanish. The model contains 35 billion parameters.\footnote{\href{https://huggingface.co/CohereForAI/c4ai-command-r-v01}{CohereForAI/c4ai-command-r-v01}} We utilize the model API available in the Cohere platform (\href{https://cohere.com/}{https://cohere.com/}), with a temperature setting of 0.1, and using the chat template format.
    \item \textbf{Command R+} is also developed keeping RAG in mind and officially supports 10 languages: English, French, Spanish, Italian, German, Brazilian Portuguese, Japanese, Korean, Arabic, and Chinese. The model contains 105 billion parameters.\footnote{\href{https://huggingface.co/CohereForAI/c4ai-command-r-plus}{CohereForAI/c4ai-command-r-plus}} We utilize the model API available in the Cohere platform (\href{https://cohere.com/}{https://cohere.com/}), with a temperature setting of 0.1, and using the chat template format.
    \item \textbf{Aya-23-35B} \cite{aryabumi:2024} is an instruction fine-tuned model with highly advanced multilingual capabilities. The model officially supports 23 languages: Arabic, Chinese (simplified \& traditional), Czech, Dutch, English, French, German, Greek, Hebrew, Hindi, Indonesian, Italian, Japanese, Korean, Persian, Polish, Portuguese, Romanian, Russian, Spanish, Turkish, Ukrainian, and Vietnamese. The model contains 35 billion parameters.\footnote{\href{https://huggingface.co/CohereForAI/aya-23-35B}{CohereForAI/aya-23-35B}} We utilize the model API available in the Cohere platform (\href{https://cohere.com/}{https://cohere.com/}), with a temperature setting of 0.1, and using the chat template format.
    \item \textbf{Gemma 1.1 (2B) it} \cite{mesnard:2024} is an instruction fine-tuned model trained using the RLHF method containing 2 billion parameters.\footnote{\href{https://huggingface.co/google/gemma-1.1-2b-it}{google/gemma-1.1-2b-it}}~We used the multiple GPU inference using the \texttt{vllm} repository \cite{kwon:2023}. We set the temperature parameter to 0.1.
    \item \textbf{Gemma 1.1 (7B) it} \cite{mesnard:2024} is an instruction fine-tuned model trained using the RLHF method containing 7 billion parameters.\footnote{\href{https://huggingface.co/google/gemma-1.1-7b-it}{google/gemma-1.1-7b-it}}~We used the multiple GPU inference using the \texttt{vllm} repository \cite{kwon:2023}. We set the temperature parameter to 0.1.
    \item \textbf{Meta-Llama-3-8B-Instruct} \cite{llama3:2024} is an English-only instruction fine-tuned model containing 8 billion parameters.\footnote{\href{https://huggingface.co/meta-llama/Meta-Llama-3-8B-Instruct}{meta-llama/Meta-Llama-3-8B-Instruct}}~We used the multiple GPU inference using the \texttt{vllm} repository \cite{kwon:2023}. We set the temperature parameter to 0.1.
    \item \textbf{Meta-Llama-3-70B-Instruct} \cite{llama3:2024} is an instruction fine-tuned model containing 70B parameters.\footnote{\href{https://huggingface.co/meta-llama/Meta-Llama-3-70B-Instruct}{meta-llama/Meta-Llama-3-70B-Instruct}}~As the model is computationally not feasible to evaluate due to resource constraints, We use the model API endpoint available in the Anyscale platform (\href{https://www.anyscale.com/}{https://www.anyscale.com/}), with a temperature setting of 0.1.
    \item \textbf{Phi-3 (mini)} \cite{phi3:2024} is an English-focused instruction fine-tuned model trained model containing 3.8 billion parameters.\footnote{\href{https://huggingface.co/microsoft/Phi-3-mini-128k-instruct}{microsoft/Phi-3-mini-128k-instruct}}~We used the multiple GPU inference using the \texttt{vllm} repository \cite{kwon:2023}. We set the temperature parameter to 0.1.
    \item \textbf{Phi-3 (small)} \cite{phi3:2024} is a multilingual instruction fine-tuned model trained model containing 8 billion parameters.\footnote{\href{https://huggingface.co/microsoft/Phi-3-small-8k-instruct}{microsoft/Phi-3-small-8k-instruct}}~There is no available information on the number of languages covered by the model. We used the multiple GPU inference using the \texttt{vllm} repository \cite{kwon:2023}. We set the temperature parameter to 0.1.
    \item \textbf{Phi-3 (medium)} \cite{phi3:2024} is a multilingual instruction fine-tuned model trained model containing 14 billion parameters.\footnote{\href{https://huggingface.co/microsoft/Phi-3-medium-128k-instruct}{microsoft/Phi-3-medium-128k-instruct}}~There is no available information on the number of languages covered by the model. We used the multiple GPU inference using the \texttt{vllm} repository \cite{kwon:2023}. We set the temperature parameter to 0.1.
    \item \textbf{Qwen2-1.5B-Instruct} \cite{qwen2:2024} is an English-focused instruction fine-tuned model trained model containing 1.5 billion parameters.\footnote{\href{https://huggingface.co/Qwen/Qwen2-1.5B-Instruct}{Qwen/Qwen2-1.5B-Instruct}}~We used the multiple GPU inference using the \texttt{vllm} repository \cite{kwon:2023}. We set the temperature parameter to 0.1.
    \item \textbf{Qwen2-7B-Instruct} \cite{qwen2:2024} is an English-focused instruction fine-tuned model trained model containing 7 billion parameters.\footnote{\href{https://huggingface.co/Qwen/Qwen2-7B-Instruct}{Qwen/Qwen2-7B-Instruct}}~We used the multiple GPU inference using the \texttt{vllm} repository \cite{kwon:2023}. We set the temperature parameter to 0.1.
\end{enumerate} 

\section{MIRAGE Fine-tuning Details}
For multilingual RAG fine-tuning, we use teacher models to \emph{distill} synthetic knowledge directly within smaller open-source models. We first generate RAG outputs on the \mirage training dataset using three high-performing teacher models: (i) GPT-4o, (ii) Llama-3 (70B), and (iii) Mixtral (8$\times$22B), and generate RAG output for queries in \mirage training dataset using only relevant passages, i.e., without distracting the model with information from non-relevant passages. We filter out the teacher model responses and curate them to create the training dataset.

Next, using supervised fine-tuning (SFT) with LoRA \cite{hu:2022}, we fine-tune two open-source models: (i) Llama-3 (8B) and (ii) Mistral-v0.2 (7B). Our hyperparameter choices are listed in \autoref{table:hyperparameters}. We use PEFT \cite{peft} and the \texttt{alignment-handbook}\footnote{\href{https://github.com/huggingface/alignment-handbook}{https://github.com/huggingface/alignment-handbook}} \cite{alignment_handbook} for supervised LoRA fine-tuning. We finetune four variants of models: (i) Mistral-v0.2 (7B) distilled using GPT-4o as a teacher, (ii)  Mistral-v0.2 (7B) distilled using Mixtral (8$\times$22B) and itself as a teacher, (iii) Llama-3 (8B) distilled using GPT-4o as a teacher, and (iv) Llama-3 (8B) distilled using Llama-3 (70B) and itself as a teacher. After fine-tuning, first all heuristic features are computed, using the already trained learning to rank model (using the baselines) is used to compute inference for the fine-tuned models and compared against upper-bound baselines, GPT-4o, Llama-3 (70B), and Mixtral (8$\times$22B) and lower-bound baselines, Mistral-v0.2 (7B) and Llama-3 (8B).

\section{Extending MIRAGE Evaluation}\label{sec:extending-eval}

As a holdout experiment, we evaluate newer versions of models, (i) Llama-3.1 series \cite{llama3:2024}: Llama-3.1 (8B)\footnote{\href{https://huggingface.co/meta-llama/Meta-Llama-3.1-8B-Instruct}{meta-llama/Meta-Llama-3.1-8B-Instruct}} and Llama-3.1 (70B)\footnote{\href{https://huggingface.co/meta-llama/Meta-Llama-3.1-70B-Instruct}{meta-llama/Meta-Llama-3.1-70B-Instruct}} instruct versions, and (ii) Gemma-2 series \cite{gemma:2024}: Gemma-2 (9B)\footnote{\href{https://huggingface.co/google/gemma-2-9b-it}{google/gemma-2-9b-it}} and Gemma-2 (27B)\footnote{\href{https://huggingface.co/google/gemma-2-27b-it}{google/gemma-2-27b-it}} instruct versions. For both models, we used the API versions of the model provided by NVIDIA (\href{https://build.nvidia.com/}{https://build.nvidia.com/}) by setting the temperature parameter to 0.1. The maximum sequence length of Gemma-2 models is 4096 tokens.

\begin{figure*}[t]
\centering
\begin{center}
    \includegraphics[trim=0 0 0 0,clip,width=0.7\textwidth]{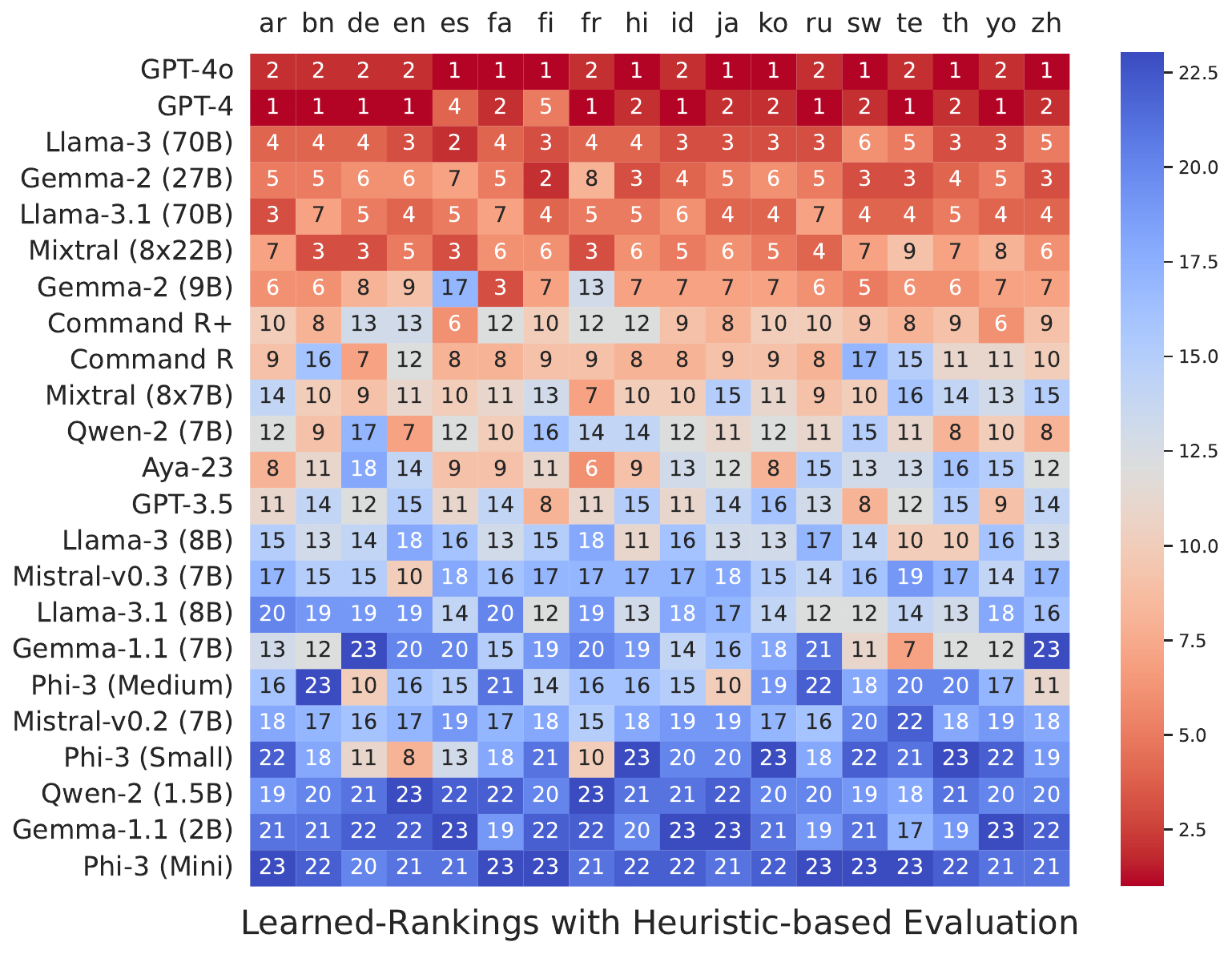}
    \caption{Approximate rankings using heuristic features including the newer models,  Llama-3.1 \cite{llama3:2024} and Gemma-2 \cite{gemma:2024} on \mirage dataset across all 18 languages. Gemma-2 (27B) and Llama-3.1 (70B) achieve a strong rank of 4 and 5 respectively in the \mirage evaluation dataset.}
    \label{fig:new-model-results}
\end{center}
\vspace*{-\baselineskip}
\end{figure*}

\smallskip
\noindent \textbf{Experimental results.} From \autoref{fig:new-model-results}, we observe that the Gemma-2 (27B) and Llama-3.1 (70B) are strong baselines, by achieving an overall rank of 4 and 5 in the \mirage dataset. Gemma-2 (27B) improves the previously best Gemma-1.1 (7B) by 13 ranks, whereas Llama-3.1 (70B) continues to underperform the best Llama-3 (70B) by 2 ranks. These results indicate newer models are improving, as reported using the surrogate judge on the synthetic \mirage leaderboard.


\begin{table*}[t]
\centering
\resizebox{0.8\textwidth}{!}{
\begin{tabular}{ccc|ccc|ccc}
\toprule
\textbf{Lang.} & \textbf{Mean} & \textbf{95\% CI} & \textbf{Lang. } & \textbf{Mean} & \textbf{95\% CI} & \textbf{Lang. } & \textbf{Mean} & \textbf{95\% CI} \\
\midrule
ar & 0.916 & -0.15 / +0.07 & bn & 0.937 & -0.17 / +0.06 & de & 0.939 & -0.14 / +0.05 \\
en & 0.971 & -0.07 / +0.03 & es & 0.844 & -0.12 / +0.09 & fa & 0.944 & -0.22 / +0.05 \\
fi & 0.957 & -0.07 / +0.04 & fr & 0.861 & -0.15 / +0.09 & hi & 0.858 & -0.26 / +0.13 \\ 
id & 0.793 & -0.17 / +0.12 & ja & 0.892 & -0.13 / +0.08 & ko & 0.941 & -0.13 / +0.06 \\ 
ru & 0.968 & -0.11 / +0.03 & sw & 0.973 & -0.06 / +0.03 & te & 0.929 & -0.16 / +0.07 \\
th & 0.902 & -0.12 / +0.09 & yo & 0.709 & -0.22 / +0.16 & zh & 0.954 & -0.09 / +0.05 \\
\bottomrule
\end{tabular}}
\caption{$\bar{R}^2$ mean scores with 95\% confidence interval with bootstrapping across all languages in \mirage. We randomly kept two models as holdout in our work: Gemma 1.1 (2B) and Llama-3 (70B).}
\label{table:r2_scores}
\end{table*}
\begin{table*}[t]
\centering
\resizebox{0.48\textwidth}{!}{
\begin{tabular}{cc}
\toprule
\textbf{Hyper-parameter} & \textbf{Choice} \\
\midrule
Attention & FlashAttention-2 \cite{dao:2024} \\
Batch Size & 32 \\
Epochs & 1 \\
Learning Rate & 2$e$-4 \\
Max Sequence Length & 6144 \\
Lora rank ($r$) & 16 \\
Lora alpha ($\alpha$) & 16 \\
Lora dropout & 0.05 \\
Lora Modules & \multicolumn{1}{p{5cm}}{[\texttt{q\_proj}, \texttt{k\_proj}, \texttt{v\_proj}, \texttt{o\_proj}, \texttt{gate\_proj}, \texttt{up\_proj}, \texttt{down\_proj}]} \\
\bottomrule
\end{tabular}}
\caption{Hyperparameter settings set during supervised fine-tuning of Mistral-v0.2 (7B) and Llama-3 (8B) on the \mirage training dataset.}
\label{table:hyperparameters}
\end{table*}

\begin{landscape}
\begin{table}[t]
    \centering
    \resizebox{1.5\textwidth}{!}{
    \begin{tabular}{lrr|lrr|lrr|lrr|lrr}
        \toprule
        \multicolumn{3}{c|}{\textbf{Arabic (ar)}} & \multicolumn{3}{c|}{\textbf{Bengali (bn)}} & \multicolumn{3}{c|}{\textbf{German (de)}} & \multicolumn{3}{c|}{\textbf{English (en)}} & \multicolumn{3}{c}{\textbf{Spanish (es)}} \\
        \textbf{Model Name} & \textbf{Mean} & \textbf{95\% CI} & \textbf{Model Name} & \textbf{Mean} & \textbf{95\% CI} & \textbf{Model Name} & \textbf{Mean} & \textbf{95\% CI} & \textbf{Model Name} & \textbf{Mean} & \textbf{95\% CI} & \textbf{Model Name} & \textbf{Mean} & \textbf{95\% CI} \\
        \midrule
    1. GPT-4 & 2.838 & -0.22/+0.28 & 1. GPT-4o & 3.104 & -0.31/+0.33 & 1. GPT-4o & 2.644 & -0.21/+0.26 & 1. GPT-4o & 2.311 & -0.22/+0.21 & 1. GPT-4o & 2.280 & -0.30/+0.32 \\
    2. GPT-4o & 2.549 & -0.27/+0.24 & 2. GPT-4 & 2.881 & -0.32/+0.34 & 2. GPT-4 & 2.498 & -0.27/+0.26 & 2. GPT-4 & 2.235 & -0.27/+0.30 & 2. GPT-4 & 1.230 & -0.28/+0.27 \\
    3. Llama-3 (70B) & 1.271 & -0.19/+0.20 & 3. Llama-3 (70B) & 1.290 & -0.28/+0.26 & 3. Mixtral (8x22B) & 1.002 & -0.21/+0.23 & 3. Llama-3 (70B) & 1.082 & -0.22/+0.19 & 3. Llama-3 (70B) & 1.225 & -0.17/+0.20 \\
    4. Mixtral (8x22B) & 1.074 & -0.23/+0.27 & 4. Mixtral (8x22B) & 1.266 & -0.22/+0.24 & 4. Llama-3 (70B) & 0.921 & -0.23/+0.22 & 4. Mixtral (8x22B) & 0.689 & -0.18/+0.18 & 4. Mixtral (8x22B) & 0.832 & -0.21/+0.21 \\
    5. Command R & 0.696 & -0.19/+0.23 & 5. Qwen2 (7B) & 0.486 & -0.20/+0.22 & 5. Mixtral (8x7B) & 0.481 & -0.25/+0.23 & 5. Qwen2 (7B) & 0.256 & -0.24/+0.24 & 5. Aya-23 & 0.485 & -0.24/+0.22 \\
    6. Command R+ & 0.476 & -0.19/+0.24 & 6. Command R+ & 0.373 & -0.18/+0.23 & 6. Command R & 0.213 & -0.21/+0.20 & 6. Phi-3 (small) & 0.198 & -0.24/+0.26 & 6. Mixtral (8x7B) & 0.476 & -0.21/+0.25 \\
    7. Aya-23 & 0.450 & -0.19/+0.20 & 7. Mixtral (8x7B) & 0.112 & -0.24/+0.22 & 7. Phi-3 (small) & 0.148 & -0.21/+0.20 & 7. Mixtral (8x7B) & 0.119 & -0.23/+0.26 & 7. Command R & 0.402 & -0.18/+0.21 \\
    8. GPT-3.5 & 0.344 & -0.28/+0.28 & 8. Aya-23 & 0.039 & -0.21/+0.19 & 8. Aya-23 & 0.057 & -0.21/+0.18 & 8. Command R & 0.011 & -0.19/+0.19 & 8. Command R+ & 0.284 & -0.20/+0.18 \\
    9. Qwen2 (7B) & 0.293 & -0.23/+0.21 & 9. Gemma-1.1 (7B) & -0.141 & -0.20/+0.21 & 9. Phi-3 (medium) & -0.027 & -0.20/+0.18 & 9. Aya-23 & -0.059 & -0.23/+0.22 & 9. GPT-3.5 & -0.029 & -0.18/+0.22 \\
    10. Mixtral (8x7B) & 0.139 & -0.24/+0.28 & 10. Llama-3 (8B) & -0.143 & -0.20/+0.19 & 10. GPT-3.5 & -0.077 & -0.18/+0.17 & 10. Command R+ & -0.193 & -0.24/+0.25 & 10. Phi-3 (small) & -0.044 & -0.21/+0.22 \\
    11. Llama-3 (8B) & -0.407 & -0.20/+0.24 & 11. GPT-3.5 & -0.388 & -0.27/+0.27 & 11. Command R+ & -0.216 & -0.20/+0.19 & 11. Mistral-v0.3 (7B) & -0.259 & -0.20/+0.23 & 11. Phi-3 (medium) & -0.056 & -0.20/+0.20 \\
    12. Gemma-1.1 (7B) & -0.444 & -0.19/+0.18 & 12. Mistral-v0.3 (7B) & -0.891 & -0.24/+0.25 & 12. Llama-3 (8B) & -0.224 & -0.16/+0.17 & 12. Mistral-v0.2 (7B)& -0.274 & -0.23/+0.23 & 12. Qwen2 (7B) & -0.095 & -0.20/+0.19 \\
    13. Phi-3 (medium) & -0.635 & -0.26/+0.31 & 13. Command R & -0.908 & -0.16/+0.15 & 13. Mistral-v0.2 (7B)& -0.331 & -0.21/+0.20 & 13. GPT-3.5 & -0.314 & -0.25/+0.25 & 13. Llama-3 (8B) & -0.247 & -0.23/+0.23 \\
    14. Mistral-v0.3 (7B) & -0.972 & -0.25/+0.23 & 14. Qwen2 (1.5B) & -1.038 & -0.28/+0.30 & 14. Qwen2 (7B) & -0.332 & -0.20/+0.23 & 14. Phi-3 (medium) & -0.330 & -0.21/+0.24 & 14. Mistral-v0.3 (7B) & -0.338 & -0.22/+0.20 \\
    15. Mistral-v0.2 (7B)& -1.076 & -0.23/+0.21 & 15. Gemma-1.1 (2B) & -1.111 & -0.27/+0.23 & 15. Mistral-v0.3 (7B) & -0.363 & -0.20/+0.20 & 15. Llama-3 (8B) & -0.444 & -0.19/+0.15 & 15. Mistral-v0.2 (7B) & -0.576 & -0.22/+0.25 \\
    16. Qwen2 (1.5B) & -1.180 & -0.25/+0.23 & 16. Mistral-v0.2 (7B)& -1.151 & -0.25/+0.22 & 16. Phi-3 (mini) & -1.262 & -0.24/+0.22 & 16. Gemma-1.1 (7B) & -0.764 & -0.19/+0.20 & 16. Gemma-1.1 (7B) & -1.023 & -0.30/+0.26 \\
    17. Gemma-1.1 (2B) & -1.611 & -0.17/+0.18 & 17. Phi-3 (small) & -1.201 & -0.19/+0.16 & 17. Qwen2 (1.5B) & -1.623 & -0.31/+0.28 & 17. Phi-3 (mini) & -1.123 & -0.23/+0.20 & 17. Phi-3 (mini) & -1.412 & -0.31/+0.29 \\
    18. Phi-3 (small) & -1.612 & -0.22/+0.24 & 18. Phi-3 (mini) & -1.269 & -0.18/+0.15 & 18. Gemma-1.1 (2B) & -1.716 & -0.23/+0.21 & 18. Gemma-1.1 (2B) & -1.308 & -0.29/+0.27 & 18. Qwen2 (1.5B) & -1.476 & -0.29/+0.28 \\
    19. Phi-3 (mini) & -2.194 & -0.23/+0.26 & 19. Phi-3 (medium) & -1.312 & -0.23/+0.20 & 19. Gemma-1.1 (7B) & -1.791 & -0.26/+0.22 & 19. Qwen2 (1.5B) & -1.832 & -0.26/+0.24 & 19. Gemma-1.1 (2B) & -1.919 & -0.32/+0.24 \\
        \bottomrule 
        \\ \\
        \toprule
        \multicolumn{3}{c|}{\textbf{Farsi (fa)}} & \multicolumn{3}{c|}{\textbf{Finnish (fi)}} & \multicolumn{3}{c|}{\textbf{French (fr)}} & \multicolumn{3}{c|}{\textbf{Hindi (hi)}} & \multicolumn{3}{c}{\textbf{Indonesian (id)}}\\
        \textbf{Model Name} & \textbf{Mean} & \textbf{95\% CI} & \textbf{Model Name} & \textbf{Mean} & \textbf{95\% CI} & \textbf{Model Name} & \textbf{Mean} & \textbf{95\% CI} & \textbf{Model Name} & \textbf{Mean} & \textbf{95\% CI} & \textbf{Model Name} & \textbf{Mean} & \textbf{95\% CI} \\
        \midrule
        1. GPT-4o & 2.951 & -0.24/+0.27 & 1. GPT-4o                              & 2.623 & -0.22/+0.24 & 1. GPT-4                               & 2.795 & -0.35/+0.40 & 1. GPT-4                               & 3.065 & -0.36/+0.31 & 1. GPT-4                               & 2.207 & -0.30/+0.28 \\
        2. GPT-4 & 2.639 & -0.27/+0.33 & 2. Llama-3 (70B)      & 1.616 & -0.20/+0.19 & 2. GPT-4o                              & 2.476 & -0.35/+0.32 & 2. GPT-4o                              & 2.958 & -0.25/+0.26 & 2. GPT-4o                              & 2.074 & -0.30/+0.34 \\
        3. Llama-3 (70B) & 1.533 & -0.17/+0.16 & 3. GPT-4                               & 1.552 & -0.30/+0.38 & 3. Llama-3 (70B)      & 1.070 & -0.22/+0.22 & 3. Llama-3 (70B)      & 1.198 & -0.21/+0.20 & 3. Llama-3 (70B)      & 1.296 & -0.21/+0.21 \\
        4. Mixtral (8x22B) & 1.257 & -0.21/+0.24 & 4. Mixtral (8x22B)     & 1.047 & -0.25/+0.24 & 4. Mixtral (8x22B)     & 1.049 & -0.24/+0.18 & 4. Mixtral (8x22B)     & 1.066 & -0.23/+0.23 & 4. Mixtral (8x22B)     & 0.848 & -0.25/+0.26 \\
        5. Command R+ & 0.748 & -0.21/+0.22 & 5. GPT-3.5                       & 0.513 & -0.20/+0.23 & 5. Aya-23                   & 0.507 & -0.22/+0.23 & 5. Command R                & 0.812 & -0.21/+0.22 & 5. Command R                & 0.337 & -0.18/+0.21 \\
        6. Command R & 0.699 & -0.19/+0.20 & 6. Command R                & 0.452 & -0.25/+0.19 & 6. Mixtral (8x7B)      & 0.504 & -0.25/+0.26 & 6. Aya-23                   & 0.646 & -0.22/+0.27 & 6. Command R+           & 0.187 & -0.16/+0.19 \\
        7. Qwen2 (7B) & 0.423 & -0.26/+0.26 & 7. Command R+           & 0.302 & -0.21/+0.19 & 7. Command R                & 0.332 & -0.18/+0.21 & 7. Command R+           & 0.586 & -0.19/+0.21 & 7. GPT-3.5                       & 0.083 & -0.22/+0.21 \\
        8. Aya-23 & 0.411 & -0.20/+0.23 & 8. Mixtral (8x7B)      & 0.242 & -0.27/+0.27 & 8. GPT-3.5                       & 0.091 & -0.20/+0.20 & 8. Mixtral (8x7B)      & 0.417 & -0.25/+0.26 & 8. Qwen2 (7B)                   & -0.007 & -0.24/+0.25 \\
        9. Mixtral (8x7B) & 0.395 & -0.26/+0.31 & 9. Aya-23                   & 0.187 & -0.21/+0.23 & 9. Command R+           & 0.026 & -0.26/+0.24 & 9. Qwen2 (7B)                    & 0.041 & -0.22/+0.21 & 9. Aya-23                  & -0.012 & -0.21/+0.20 \\
        10. Llama-3 (8B) & -0.024 & -0.26/+0.25 & 10. Phi-3 (medium)     & -0.068 & -0.26/+0.23 & 10. Phi-3 (small)        & -0.058 & -0.24/+0.26 & 10. Llama-3 (8B)      & -0.032 & -0.20/+0.25 & 10. Mixtral (8x7B)     & -0.046 & -0.27/+0.26 \\
        11. GPT-3.5 & -0.091 & -0.23/+0.27 & 11. Llama-3 (8B)      & -0.088 & -0.19/+0.17 & 11. Qwen2 (7B)                   & -0.175 & -0.27/+0.27 & 11. GPT-3.5                      & -0.170 & -0.22/+0.24 & 11. Llama-3 (8B)      & -0.274 & -0.17/+0.19 \\
        12. Gemma-1.1 (7B) & -0.671 & -0.22/+0.20 & 12. Qwen2 (7B)                   & -0.246 & -0.24/+0.22 & 12. Phi-3 (medium)     & -0.385 & -0.23/+0.19 & 12. Mistral-v0.3 (7B)       & -0.519 & -0.22/+0.28 & 12. Mistral-v0.3 (7B)       & -0.321 & -0.22/+0.27 \\
        13. Mistral-v0.3 (7B) & -0.763 & -0.27/+0.23 & 13. Mistral-v0.3 (7B)       & -0.543 & -0.17/+0.19 & 13. Mistral-v0.2 (7B)       & -0.442 & -0.21/+0.22 & 13. Phi-3 (medium)     & -0.620 & -0.31/+0.28 & 13. Gemma-1.1 (7B)                   & -0.466 & -0.23/+0.21 \\
        14. Mistral-v0.2 (7B) & -1.127 & -0.22/+0.23 & 14. Mistral-v0.2 (7B)       & -0.756 & -0.17/+0.20 & 14. Mistral-v0.3 (7B)       & -0.457 & -0.22/+0.21 & 14. Mistral-v0.2 (7B)       & -0.873 & -0.18/+0.19 & 14. Phi-3 (medium)     & -0.529 & -0.21/+0.23 \\
        15. Gemma-1.1 (2B) & -1.258 & -0.23/+0.18 & 15. Qwen2 (1.5B)                 & -0.975 & -0.22/+0.24 & 15. Llama-3 (8B)      & -0.653 & -0.17/+0.18 & 15. Gemma-1.1 (2B)                   & -1.205 & -0.27/+0.23 & 15. Mistral-v0.2 (7B)       & -0.591 & -0.25/+0.19 \\
        16. Phi-3 (medium) & -1.368 & -0.21/+0.24 & 16. Gemma-1.1 (7B)                   & -1.029 & -0.19/+0.20 & 16. Phi-3 (mini)       & -1.353 & -0.27/+0.27 & 16. Gemma-1.1 (7B)                   & -1.257 & -0.25/+0.27 & 16. Phi-3 (small)        & -0.627 & -0.26/+0.25 \\
        17. Qwen2 (1.5B) & -1.501 & -0.21/+0.26 & 17. Phi-3 (small)        & -1.316 & -0.23/+0.23 & 17. Gemma-1.1 (7B)                   & -1.491 & -0.24/+0.25 & 17. Qwen2 (1.5B)                 & -1.529 & -0.25/+0.25 & 17. Qwen2 (1.5B)                 & -0.967 & -0.27/+0.26 \\
        18. Phi-3 (small) & -1.623 & -0.21/+0.18 & 18. Gemma-1.1 (2B)                   & -1.420 & -0.24/+0.23 & 18. Qwen2 (1.5B)                 & -1.861 & -0.32/+0.28 & 18. Phi-3 (mini)       & -2.192 & -0.20/+0.22 & 18. Phi-3 (mini)       & -1.514 & -0.29/+0.26 \\
        19. Phi-3 (mini) & -2.629 & -0.34/+0.28 & 19. Phi-3 (mini)       & -2.092 & -0.26/+0.26 & 19. Gemma-1.1 (2B)                   & -1.977 & -0.24/+0.25 & 19. Phi-3 (small)        & -2.394 & -0.31/+0.27 & 19. Gemma-1.1 (2B)                   & -1.678 & -0.25/+0.26 \\
     \bottomrule
    \end{tabular}}
    \caption{Bradley-Terry logits using GPT-4o as a judge for all languages in \mirage. Scores are computed using the Bradley-Terry model with 200 tournaments using a maximum of 100 randomly sampled queries. Mean scores and 95\% confidence intervals are reported (repeated 200 times). A higher logit score indicates a better performance, therefore achieving a higher rank on \mirage. Models are sorted in descending order of mean score for each language in \mirage.}
    \label{table:bradley_terry_page1}
    \end{table}
\end{landscape}

\begin{landscape}
\begin{table}[t]
    \centering
    \resizebox{1.2\textwidth}{!}{
    \begin{tabular}{lrr|lrr|lrr|lrr}
        \toprule
        \multicolumn{3}{c|}{\textbf{Japanesee (ja)}} & \multicolumn{3}{c|}{\textbf{Korean (ko)}} & \multicolumn{3}{c|}{\textbf{Russian (ru)}} & \multicolumn{3}{c}{\textbf{Swahili (sw)}} \\
        \textbf{Model Name} & \textbf{Mean} & \textbf{95\% CI} & \textbf{Model Name} & \textbf{Mean} & \textbf{95\% CI} & \textbf{Model Name} & \textbf{Mean} & \textbf{95\% CI} & \textbf{Model Name} & \textbf{Mean} & \textbf{95\% CI}\\
        \midrule
    1. GPT-4o                              & 2.294 & -0.24/+0.32 & 1. GPT-4o                              & 2.402 & -0.34/+0.35 & 1. GPT-4                               & 2.372 & -0.26/+0.25 & 1. GPT-4o                              & 2.966 & -0.25/+0.28 \\
    2. GPT-4                               & 2.173 & -0.24/+0.24 & 2. GPT-4                               & 2.158 & -0.26/+0.31 & 2. GPT-4o                              & 2.208 & -0.27/+0.25 & 2. GPT-4                               & 2.038 & -0.36/+0.40 \\
    3. Llama-3 (70B)      & 1.088 & -0.22/+0.23 & 3. Llama-3 (70B)      & 1.391 & -0.23/+0.21 & 3. Llama-3 (70B)      & 1.323 & -0.20/+0.21 & 3. Llama-3 (70B)      & 1.750 & -0.22/+0.20 \\
    4. Mixtral (8x22B)     & 0.834 & -0.18/+0.19 & 4. Mixtral (8x22B)     & 1.028 & -0.24/+0.23 & 4. Mixtral (8x22B)     & 1.196 & -0.24/+0.25 & 4. Mixtral (8x22B)     & 1.355 & -0.19/+0.22 \\
    5. Command R                & 0.377 & -0.20/+0.20 & 5. Command R                & 0.647 & -0.20/+0.26 & 5. Mixtral (8x7B)      & 0.821 & -0.26/+0.27 & 5. GPT-3.5                       & 1.006 & -0.24/+0.24 \\
    6. Aya-23                   & 0.290 & -0.21/+0.19 & 6. Aya-23                   & 0.645 & -0.20/+0.20 & 6. Command R                & 0.687 & -0.16/+0.17 & 6. Command R+           & 0.735 & -0.19/+0.17 \\
    7. Qwen2 (7B)                    & 0.269 & -0.27/+0.27 & 7. Mixtral (8x7B)      & 0.617 & -0.25/+0.20 & 7. Aya-23                   & 0.276 & -0.17/+0.20 & 7. Mixtral (8x7B)      & 0.473 & -0.23/+0.24 \\
    8. Phi-3 (medium)      & 0.225 & -0.21/+0.25 & 8. Qwen2 (7B)                    & 0.218 & -0.25/+0.23 & 8. GPT-3.5                       & 0.261 & -0.25/+0.21 & 8. Gemma-1.1 (7B)                    & 0.385 & -0.23/+0.23 \\
    9. Mixtral (8x7B)     & -0.020 & -0.27/+0.27 & 9. Command R+           & 0.026 & -0.18/+0.17 & 9. Command R+           & 0.121 & -0.19/+0.20 & 9. Aya-23                   & -0.042 & -0.24/+0.24 \\
    10. Command R+          & -0.044 & -0.17/+0.21 & 10. Llama-3 (8B)      & -0.041 & -0.20/+0.20 & 10. Qwen2 (7B)                   & 0.116 & -0.28/+0.26 & 10. Llama-3 (8B)      & -0.057 & -0.23/+0.20 \\
    11. Llama-3 (8B)      & -0.099 & -0.18/+0.18 & 11. Mistral-v0.3 (7B)       & -0.321 & -0.24/+0.24 & 11. Mistral-v0.3 (7B)       & -0.074 & -0.24/+0.24 & 11. Qwen2 (7B)                   & -0.175 & -0.30/+0.25 \\
    12. GPT-3.5                      & -0.292 & -0.21/+0.20 & 12. GPT-3.5                      & -0.352 & -0.21/+0.23 & 12. Mistral-v0.2 (7B)       & -0.187 & -0.28/+0.23 & 12. Phi-3 (medium)     & -0.783 & -0.25/+0.27 \\
    13. Gemma-1.1 (7B)                   & -0.418 & -0.22/+0.22 & 13. Mistral-v0.2 (7B)       & -0.467 & -0.27/+0.27 & 13. Llama-3 (8B)      & -0.401 & -0.19/+0.21 & 13. Mistral-v0.3 (7B)       & -0.804 & -0.27/+0.25 \\
    14. Mistral-v0.3 (7B)       & -0.498 & -0.24/+0.25 & 14. Gemma-1.1 (7B)                   & -0.513 & -0.23/+0.26 & 14. Phi-3 (small)        & -1.026 & -0.24/+0.28 & 14. Command R               & -1.008 & -0.23/+0.23 \\
    15. Mistral-v0.2 (7B)       & -0.599 & -0.22/+0.25 & 15. Phi-3 (medium)     & -1.046 & -0.24/+0.27 & 15. Qwen2 (1.5B)                 & -1.135 & -0.26/+0.22 & 15. Qwen2 (1.5B)                 & -1.077 & -0.26/+0.27 \\
    16. Phi-3 (small)        & -0.947 & -0.30/+0.27 & 16. Qwen2 (1.5B)                 & -1.162 & -0.27/+0.22 & 16. Gemma-1.1 (2B)                   & -1.216 & -0.23/+0.26 & 16. Mistral-v0.2 (7B)       & -1.356 & -0.29/+0.24 \\
    17. Qwen2 (1.5B)                 & -1.471 & -0.27/+0.26 & 17. Gemma-1.1 (2B)                   & -1.263 & -0.25/+0.27 & 17. Gemma-1.1 (7B)                   & -1.469 & -0.21/+0.23 & 17. Gemma-1.1 (2B)                   & -1.448 & -0.27/+0.23 \\
    18. Phi-3 (mini)       & -1.548 & -0.26/+0.24 & 18. Phi-3 (small)        & -1.864 & -0.24/+0.23 & 18. Phi-3 (medium)     & -1.758 & -0.17/+0.18 & 18. Phi-3 (small)        & -1.937 & -0.30/+0.35 \\
    19. Gemma-1.1 (2B)                   & -1.615 & -0.32/+0.23 & 19. Phi-3 (mini)       & -2.104 & -0.22/+0.24 & 19. Phi-3 (mini)       & -2.115 & -0.25/+0.25 & 19. Phi-3 (mini)       & -2.019 & -0.30/+0.29 \\
        \bottomrule
        \\ \\
        \toprule
        \multicolumn{3}{c|}{\textbf{Telugu (te)}} & \multicolumn{3}{c|}{\textbf{Thai (th)}} & \multicolumn{3}{c|}{\textbf{Yoruba (yo)}} & \multicolumn{3}{c}{\textbf{Chinese (zh)}} \\
        \textbf{Model Name} & \textbf{Mean} & \textbf{95\% CI} & \textbf{Model Name} & \textbf{Mean} & \textbf{95\% CI} & \textbf{Model Name} & \textbf{Mean} & \textbf{95\% CI} & \textbf{Model Name} & \textbf{Mean} & \textbf{95\% CI} \\
        \midrule
    1. GPT-4                               & 3.112 & -0.34/+0.33 & 1. GPT-4o                              & 2.606 & -0.27/+0.27 & 1. GPT-4o                              & 2.606 & -0.27/+0.27 & 1. GPT-4o                              & 2.352 & -0.25/+0.27 \\
    2. GPT-4o                              & 3.083 & -0.26/+0.32 & 2. GPT-4                               & 2.348 & -0.27/+0.30 & 2. GPT-4                               & 2.348 & -0.27/+0.30 & 2. GPT-4                               & 2.173 & -0.31/+0.31 \\
    3. Llama-3 (70B)      & 1.229 & -0.23/+0.22 & 3. Llama-3 (70B)      & 1.405 & -0.20/+0.21 & 3. Llama-3 (70B)      & 1.405 & -0.20/+0.21 & 3. Mixtral (8x22B)     & 0.987 & -0.23/+0.24 \\
    4. Gemma-1.1 (7B)                    & 1.046 & -0.26/+0.21 & 4. Mixtral (8x22B)     & 1.083 & -0.28/+0.25 & 4. Mixtral (8x22B)     & 1.083 & -0.28/+0.25 & 4. Llama-3 (70B)      & 0.950 & -0.20/+0.20 \\
    5. Command R+           & 0.566 & -0.21/+0.23 & 5. Qwen2 (7B)                    & 0.855 & -0.26/+0.27 & 5. Qwen2 (7B)                    & 0.855 & -0.26/+0.27 & 5. Command R                & 0.502 & -0.20/+0.20 \\
    6. Mixtral (8x22B)     & 0.466 & -0.22/+0.25 & 6. Command R+           & 0.513 & -0.22/+0.19 & 6. Command R+           & 0.513 & -0.22/+0.19 & 6. Qwen2 (7B)                    & 0.455 & -0.26/+0.26 \\
    7. Llama-3 (8B)       & 0.255 & -0.19/+0.21 & 7. Llama-3 (8B)       & 0.421 & -0.28/+0.24 & 7. Llama-3 (8B)       & 0.421 & -0.28/+0.24 & 7. Command R+           & 0.344 & -0.21/+0.19 \\
    8. Qwen2 (7B)                    & 0.111 & -0.26/+0.27 & 8. Gemma-1.1 (7B)                    & -0.045 & -0.19/+0.23 & 8. Gemma-1.1 (7B)                    & -0.045 & -0.19/+0.23 & 8. Phi-3 (medium)      & 0.125 & -0.20/+0.19 \\
    9. GPT-3.5                       & 0.074 & -0.23/+0.25 & 9. Command R                & -0.135 & -0.19/+0.20 & 9. Command R                & -0.135 & -0.19/+0.20 & 9. Aya-23                   & 0.106 & -0.22/+0.22 \\
    10. Aya-23                  & -0.203 & -0.24/+0.22 & 10. Mixtral (8x7B)     & -0.201 & -0.28/+0.28 & 10. Mixtral (8x7B)     & -0.201 & -0.28/+0.28 & 10. GPT-3.5                      & -0.163 & -0.23/+0.21 \\
    11. Command R               & -0.404 & -0.26/+0.23 & 11. GPT-3.5                      & -0.233 & -0.30/+0.28 & 11. GPT-3.5                      & -0.233 & -0.30/+0.28 & 11. Mixtral (8x7B)     & -0.177 & -0.23/+0.21 \\
    12. Mixtral (8x7B)     & -0.739 & -0.25/+0.28 & 12. Mistral-v0.3 (7B)       & -0.332 & -0.23/+0.31 & 12. Mistral-v0.3 (7B)       & -0.332 & -0.23/+0.31 & 12. Llama-3 (8B)      & -0.200 & -0.20/+0.22 \\
    13. Gemma-1.1 (2B)                   & -1.045 & -0.20/+0.24 & 13. Aya-23                  & -0.365 & -0.22/+0.19 & 13. Aya-23                  & -0.365 & -0.22/+0.19 & 13. Mistral-v0.3 (7B)       & -0.261 & -0.25/+0.25 \\
    14. Phi-3 (medium)     & -1.110 & -0.20/+0.20 & 14. Mistral-v0.2 (7B)       & -1.008 & -0.25/+0.25 & 14. Mistral-v0.2 (7B)       & -1.008 & -0.25/+0.25 & 14. Mistral-v0.2 (7B)       & -0.584 & -0.21/+0.20 \\
    15. Qwen2 (1.5B)                 & -1.119 & -0.20/+0.16 & 15. Phi-3 (medium)     & -1.040 & -0.22/+0.19 & 15. Phi-3 (medium)     & -1.040 & -0.22/+0.19 & 15. Phi-3 (small)        & -0.883 & -0.24/+0.26 \\
    16. Mistral-v0.3 (7B)       & -1.194 & -0.19/+0.18 & 16. Gemma-1.1 (2B)                   & -1.103 & -0.24/+0.22 & 16. Gemma-1.1 (2B)                   & -1.103 & -0.24/+0.22 & 16. Qwen2 (1.5B)                 & -0.919 & -0.22/+0.24 \\
    17. Phi-3 (small)        & -1.275 & -0.18/+0.18 & 17. Qwen2 (1.5B)                 & -1.255 & -0.38/+0.32 & 17. Qwen2 (1.5B)                 & -1.255 & -0.38/+0.32 & 17. Gemma-1.1 (2B)                   & -1.265 & -0.26/+0.31 \\
    18. Mistral-v0.2 (7B)       & -1.348 & -0.20/+0.18 & 18. Phi-3 (small)        & -1.740 & -0.22/+0.22 & 18. Phi-3 (small)        & -1.740 & -0.22/+0.22 & 18. Phi-3 (mini)       & -1.503 & -0.26/+0.28 \\
    19. Phi-3 (mini)       & -1.504 & -0.26/+0.19 & 19. Phi-3 (mini)       & -1.774 & -0.24/+0.22 & 19. Phi-3 (mini)       & -1.774 & -0.24/+0.22 & 19. Gemma-1.1 (7B)                   & -2.037 & -0.38/+0.32 \\
     \bottomrule
    \end{tabular}}
    \caption{Bradley-Terry logits using GPT-4o as a judge for all languages in \mirage. Scores are computed using the Bradley-Terry model with 200 tournaments using a maximum of 100 randomly sampled queries. Mean scores and 95\% confidence intervals are reported (repeated 200 times). A higher logit score indicates a better performance, therefore achieving a higher rank on \mirage. Models are sorted in descending order of mean score for every language in \mirage.}
    \label{table:bradley_terry_page2}
    \end{table}
\end{landscape}

\begin{figure*}[t]
\centering
\begin{center}
    \includegraphics[trim=0 0 0 0,clip,width=\textwidth]{feature-plots/language-detection-en.pdf}
    \includegraphics[trim=0 0 0 0,clip,width=\textwidth]{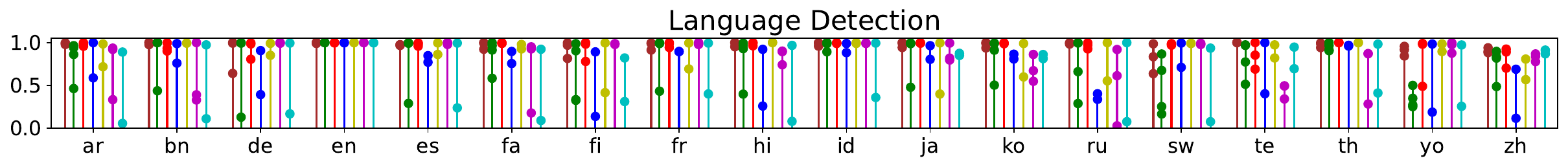}
    \includegraphics[trim=0 0 0 0,clip,width=\textwidth]{feature-plots/citation-quality-recall-10.pdf} 
    \includegraphics[trim=0 0 0 0,clip,width=\textwidth]{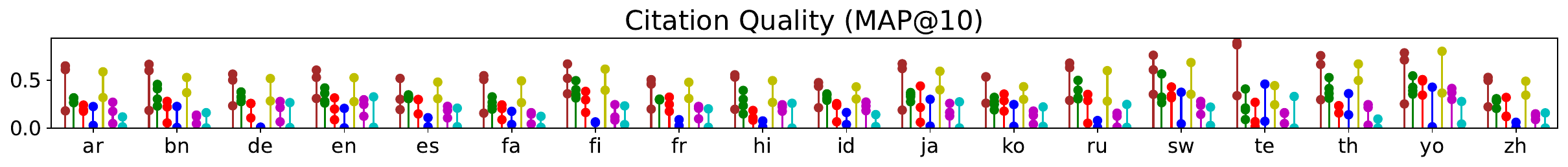} 
    \includegraphics[trim=0 0 0 0,clip,width=\textwidth]{feature-plots/support-xlni-context-grounding.pdf}
    \includegraphics[trim=0 0 0 0,clip,width=\textwidth]{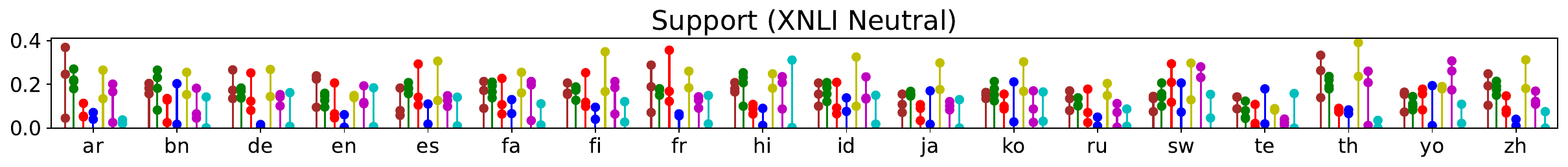}
    \includegraphics[trim=0 0 0 0,clip,width=\textwidth]{feature-plots/reranker-score-bge-v2-m3.pdf}
    \includegraphics[trim=0 0 0 0,clip,width=\textwidth]{feature-plots/answer-overlap-rouge-l.pdf}
    \includegraphics[trim=0 0 0 0,clip,width=\textwidth]{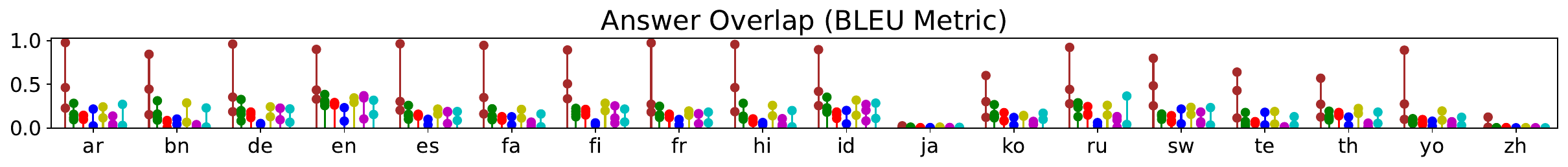}
    \includegraphics[trim=0 0 0 0,clip,width=\textwidth]{feature-plots/llm-answer-overlap-feature.pdf}
    \includegraphics[trim=0 0 0 0,clip,width=\textwidth]{feature-plots/llm-fluency-feature.pdf}
    \caption{Lollipop plots denoting the average heuristic-based feature scores achieved by baselines in \mirage for all eleven heurisitc-based features. $x$-axis denotes the languages in \mirage; whereas $y$-axis plots every heuristic feature value. Multiple LLMs in the same family are represented as a single color lollipop (multiple circles).}
    \label{fig:heuristics-appendix}
\end{center}
\vspace*{-\baselineskip}
\end{figure*}

\begin{figure*}[t!]
\begin{mdframed}[backgroundcolor=gray!5]
    \small
    \texttt{\textbf{Question:} \\
    What was the first newspaper ever printed in the U.K.? \\\\
    \textbf{Contexts:} \\ "[36897421\#2]" Lögberg-Heimskringla - The very first newspaper to be published in North America by the Icelandic immigrant population was handwritten by Jon Gudmundsson in 1876 ... \\
    "[1965416\#2]" The New York Times Magazine - Its first issue was published on September 6, 1896, and contained the first photographs ever printed in the newspaper...  \\ 
    "[662134\#6]" Letterpress printing - Letterpress printing was introduced in Canada in 1752 in Halifax, Nova Scotia by John Bushell in the newspaper format. This paper was named the Halifax Gazette and became Canada's first newspaper ... \\ \\ ... \\ ... \\  \\
    "[22112840\#15]" Newspaper - The emergence of the new media in the 17th century has to be seen in close connection with the spread of the printing press from which the publishing press derives its name.... \\\\
    \textbf{Instruction:} \\
    Provide an answer to the question using the information provided in contexts written in \{\{language\}\}. Additionally, provide a step-by-step explanation of your reasoning, demonstrating how you arrived at your answer in \{\{language\}\}. Cite parts of your reasoning within brackets [] using the IEEE format based on the provided contexts. \\ Please respond in \{\{language\}\} using the format: \#\#Reason: \{reason\} \#\#Answer: \{answer\}.
}
\end{mdframed}
\caption{Prompt template for all baseline models for multilingual RAG generation for queries across all languages in \mirage. We include the language-specific query in \mirage under ``Question:''. Next, we concatenate both relevant and non-relevant passages (randomly shuffled and truncated at maximum length) and place them under ``Contexts:''. Lastly, we provide our instruction in English asking the model to generate a response in the required language under the placeholder ``\{\{language\}\}''. The example above is shown for a query in English (en) from \mirage, where contexts are truncated ( ... ) for demonstration purposes.}
\label{fig:prompt_baseline}
\end{figure*}

\begin{figure*}[t!]
\begin{mdframed}[backgroundcolor=gray!5]
    \small
    \texttt{You are an AI assistant. In the following task, you are given a Question, a RAG application's response, and a Ground-truth Answer referred to as 'Label' in \{\{language\}\}. Assess how well the RAG application's response aligns with the Label, using the grading rubric below: \\\\
    1: The response is not aligned with the Label or is off-topic; it includes hallucination. \\
    2: The response admits it cannot provide an answer or lacks context; it is honest. \\
    3: The response is relevant but contains notable discrepancies or inaccuracies. \\
    4: The response is acceptable and sufficient but not exhaustive. \\
    5: The response is fully accurate and comprehensive, based on the Label. \\\\
    Treat the Label as the definitive answer. Present your final score in the format: "[[score]]",
    followed by your justification in English. Example: \\
    Score: [[3]] Justification: The response partially aligns with the Label but with some discrepancies.\\\\
    Question in \{\{language\}\}: \\
    \{\{Question\}\} \\\\
    Label in \{\{language\}\}: \\
    \{\{Label\}\} \\\\
    RAG Application Response in \{\{language\}\}: \\
    \{\{Response\}\} \\\\
    Treat the Label as the definitive answer. Present your final score in the format: "[[score]]",
    followed by your justification in English.
}
\end{mdframed}
\caption{Prompt template used by Llama-3 (8B) model as a judge to evaluate the answer overlap heuristic feature. We include a grading rubric within the prompt template.\{\{Label\}\} is a placeholder for the gold truth answer provided using the GPT-4; \{\{language\}\} is a placeholder for the target language; \{\{Question\}\} is a placeholder for the \mirage query; \{\{Documents\}\} is a placeholder for both \mirage relevant and non-relevant passages concatenated together; \{\{Response\}\} is a placeholder for RAG model output.}
\label{fig:prompt_answer}
\end{figure*}

\begin{figure*}[t!]
\begin{mdframed}[backgroundcolor=gray!5]
    \small
    \texttt{You will be given one summary written for a question and documents from Wikipedia in \{\{language\}\}. Your task is to rate the summary on one metric. Please make sure you read and understand these instructions carefully. Please keep this document open while reviewing, and refer to it as needed. \\\\
    Evaluation Criteria: \\
Fluency (1-5) - the collective quality of all sentences. We align this dimension with
the DUC quality question of structure and fluency whereby ``the summary should be
well-structured and well-organized. The summary should not just be a heap of related information, but should build from sentence to sentence to a coherent body of information about a topic.'' \\\\
Evaluation Steps: \\
1. Read the question and Wikipedia documents in \{\{language\}\} carefully and identify the main topic and key points. \\
2. Read the summary and check whether it answers the question. Check if the summary covers the main
topic and key points required to answer the question, and if it presents them in a clear and logical order. \\
3. Assign a rating for fluency on a scale of 1 to 5 and provide an explanation, where 1 is the lowest and 5 is the highest
based on the Evaluation Criteria. \\\\
Example: \\
Question in \{\{language\}\}: \\
\{\{Question\}\} \\\\
Documents in \{\{language\}\}: \\
\{\{Documents\}\} \\\\
Summary: \\
\{\{Summary\}\} \\\\
Rate the fluency of the summary on a scale of 1 to 5 and explain your rating. Please use the format of: \#\#Rating: \{rating\} \#\#Explanation: \{explanation\}.
}
\end{mdframed}
\caption{Prompt template used by Llama-3 (8B) model as a judge to evaluate the fluency of a RAG response. We first explain the criteria for evaluation and the model outputs an explanation and score between [1,5] indicating the fluency of the output. \{\{language\}\} is a placeholder for the target language; \{\{Question\}\} is a placeholder for the \mirage query; \{\{Documents\}\} is a placeholder for both \mirage relevant and non-relevant passages concatenated together; \{\{Summary\}\} is a placeholder for RAG model output.}
\label{fig:prompt_fluency}
\end{figure*}

\begin{figure*}[t!]
\begin{mdframed}[backgroundcolor=gray!5]
    \small
    \texttt{Please act as an impartial judge and evaluate the quality of the responses provided
by two AI assistants tasked to answer the question displayed below, based on a set
of documents retrieved by a search engine. \\
You should choose the assistant that best answers the user question based on a set
of reference documents that may or not be relevant referenced in the IEEE format. \\\\
Your evaluation should consider factors such as the correctness, helpfulness, completeness, accuracy, depth, and level of detail of their responses. \\\\
Details are only useful if they answer the user's question. If an answer 
contains non-relevant details, it should not be preferred over one that only 
uses relevant information. \\\\
Begin your evaluation by explaining why each answer correctly answers the user 
question. Then, you should compare the two responses and provide a short explanation 
on their differences. Avoid any position biases and ensure that the order in which 
the responses were presented does not influence your decision. Do not allow the 
length of the responses to influence your evaluation. Be as objective as possible. \\\\
After providing your explanation, output your final verdict by strictly following 
this format: "[[A]]" if assistant A is better, "[[B]]" if assistant B is better,
and "[[C]]" for a tie. \\\\
"[User Question]" \\
\{\{query\}\} \\\\
"[Reference Documents]" \\
\{\{documents\}\} \\\\
"[The Start of Assistant A's Answer]" \\
\{\{answer\_a\}\} \\
"[The End of Assistant A's Answer]" \\\\
"[The Start of Assistant B's Answer]" \\
\{\{answer\_b\}\} \\
"[The End of Assistant B's Answer]"
}
\end{mdframed}
\caption{Prompt template used by LLM as a judge to evaluate the RAG response in a pairwise evaluation involving a head-to-head battle. The template is taken and modified from RAGEval \cite{rackauckas:2024}. We explain the evaluation criteria and ask the judge to evaluate two RAG responses based on multiple factors, including correctness, helpfulness, completeness, accuracy, depth, and level of detail. The Judge provides a justification for their model choice and at the end of the response indicates as either ``[[A]]'', ``[[B]]'', or ``[[C]]'' denoting a tie. \{\{query\}\} is a placeholder for the input \mirage query; \{\{documents\}\} is a placeholder for both \mirage relevant and non-relevant passages concatenated together; \{\{answer\_a\}\} is a placeholder for the output response of model A; \{\{answer\_b\}\} is a placeholder for the output response of model B.}
\label{fig:prompt_gpt4}
\end{figure*}

\end{document}